%% file: crop.tex
\documentclass[letterpaper]{article} 
\usepackage{aaai24}  
\usepackage{times}  
\usepackage{helvet}  
\usepackage{courier}  
\usepackage[hyphens]{url}  
\usepackage{graphicx} 
\usepackage{natbib}  
\usepackage{caption} 
\usepackage{algorithm}
\usepackage{algorithmic}

\usepackage{newfloat}
\usepackage{listings}
\DeclareCaptionStyle{ruled}{labelfont=normalfont,labelsep=colon,strut=off} 
\floatstyle{ruled}
\newfloat{listing}{tb}{lst}{}
\floatname{listing}{Listing}
\usepackage{amsmath}
\usepackage{amssymb}
\usepackage{amsfonts}
\usepackage{tabularx}
\usepackage{booktabs}
\usepackage{subcaption}
\usepackage{xspace}

\title{Learning Subject-Aware Cropping by Outpainting Professional Photos}
\author{
    James Hong\textsuperscript{\rm 1},
	Lu Yuan\textsuperscript{\rm 1},
	Micha\"el Gharbi\textsuperscript{\rm 2},
	Matthew Fisher\textsuperscript{\rm 2},
	Kayvon Fatahalian\textsuperscript{\rm 1}
}
\affiliations{
	\textsuperscript{\rm 1}Stanford University \\
	\textsuperscript{\rm 2}Adobe Research \\
	\{ james.hong , luyuan , kayvonf \}@cs.stanford.edu,
	\{ mgharbi , matfishe \}@adobe.com
}
\nocopyright

\input{src/notation.tex}

\begin{document}

\maketitle

\input{src/abstract.tex}

\input{src/intro.tex}
\input{src/related.tex}
\input{src/method.tex}
\input{src/dataset.tex}
\input{src/result.tex}
\input{src/discussion.tex}

\input{src/ack.tex}
\input{src/ethics.tex}

\bibliography{crop}

\newpage
\appendix
\input{src_supp/supp_result.tex}

\input{src_supp/supp_dataset.tex}
\input{src_supp/supp_impl.tex}

\end{document}

%% file: src/notation.tex
\newcommand{\OURMETHOD}{{GenCrop}\xspace}

\newcommand{\fromhcic}{\textsuperscript{*}}
\newcommand{\fromsacd}{\textsuperscript{\textdagger}}

\newcommand{\best}[1]{\textbf{#1}}

%% file: src/abstract.tex
\begin{abstract}
How to frame (or crop) a photo often depends on the image subject and its context; e.g., a human portrait.
Recent works have defined the subject-aware image cropping task as a nuanced and practical version of image cropping.
We propose a weakly-supervised approach (\OURMETHOD) to learn what makes a high-quality, subject-aware crop from professional stock images.
Unlike supervised prior work, \OURMETHOD requires no new manual annotations beyond the existing stock image collection.
The key challenge in learning from this data, however, is that the images are already cropped and we do not know what regions were removed.
Our insight is to combine a library of stock images with a modern, pre-trained text-to-image diffusion model.
The stock image collection provides diversity, and its images serve as pseudo-labels for a good crop. The text-image diffusion model is used to out-paint (i.e., outward inpainting) realistic uncropped images.
Using this procedure, we are able to automatically generate a large dataset of cropped-uncropped training pairs to train a cropping model.
Despite being weakly-supervised, \OURMETHOD is competitive with state-of-the-art supervised methods and significantly better than comparable weakly-supervised baselines on quantitative and qualitative evaluation metrics.
\end{abstract}

%% file: src/intro.tex
\section{Introduction}

Framing a photo is compositional skill that professional photographers hone with years of experience, and cropping is a key way to adjust framing or experiment with alternative compositions after capture.
Framing and cropping decide what elements of a scene to include or exclude from the image, and how to frame or crop an image is influenced by the subject that one wishes to portray.
Subject-aware cropping takes a notion of a subject in addition to pixels and has been studied in recent work on data-driven approaches~\cite{sacd} and in the context of human-centric images~\cite{hcic}.
High-quality solutions to this problem are based on supervised learning, from large datasets of manual annotations created specifically for cropping~\cite{gaic,cpc,sacd}.

We explore an alternative approach to the subject-aware cropping problem that is only \emph{weakly-supervised}.
We observe that millions of professional images are easily accessible online in the form of stock image collections and that these collections cover a wide range of subject-matter that people want to capture --- e.g., portraits of people.
We then ask, {\em to crop better portraits, can one seek out a relevant set of professional portraits from these collections and teach a model to replicate that distribution?}
The key challenge is that, although every professional image provides an expert label (i.e., a good crop), the original uncropped photo is unknown and cannot be recovered from the crop.
Unlike typical weakly-supervised computer vision tasks where input images are plentiful but labels are scarce and/or unreliable, in our setting this assumption is reversed.

Our proposed method, \OURMETHOD, addresses this challenge by combining a readily available dataset of stock images with powerful, pre-trained image generation models to synthesize the required inputs.
Specifically, we use text-to-image diffusion to ``out-paint'' (i.e., outward pixel generate or outward inpaint) stock images and generate plausible uncropped-and-cropped pairs (Fig.~\ref{fig:teaser}).
By scaling this automatic process, we can generate a large and diverse set of images to train our subject-aware cropping model.
The key advantage of \OURMETHOD is that it is weakly-supervised, requiring no new manual crop or scoring annotations beyond access to the original professional image collection.

\input{src/figure/teaser}

To demonstrate the effectiveness of \OURMETHOD, we evaluate it on the human-centric (portrait) and the subject-aware cropping tasks proposed by~\cite{hcic} and~\cite{sacd}.
We show that \OURMETHOD yields competitive results against fully supervised approaches~\cite{hcic} on the existing datasets~\cite{flms,fcdb,sacd} (under quantitative metrics such as Intersection-over-Union and boundary displacement~\cite{hcic}), while being superior to the best weakly/unsupervised method~\cite{vfn}.
We also evaluate \OURMETHOD on additional subject categories such as cats, dogs, etc. to test the generalization of our approach beyond just humans.
On qualitative evaluation, \OURMETHOD is comparable to or better than supervised prior work on the rate of cropping errors, while prior weakly-supervised/unsupervised baselines fall substantially short.

Lastly, we conduct additional analysis and ablations to assess the effectiveness and limitations of learning to crop from our generated data.
The code for our data generation pipeline and cropping models is publicly available.

%% file: src/figure/teaser.tex
\begin{figure}[tp]
    \centering
    \includegraphics[width=\columnwidth]{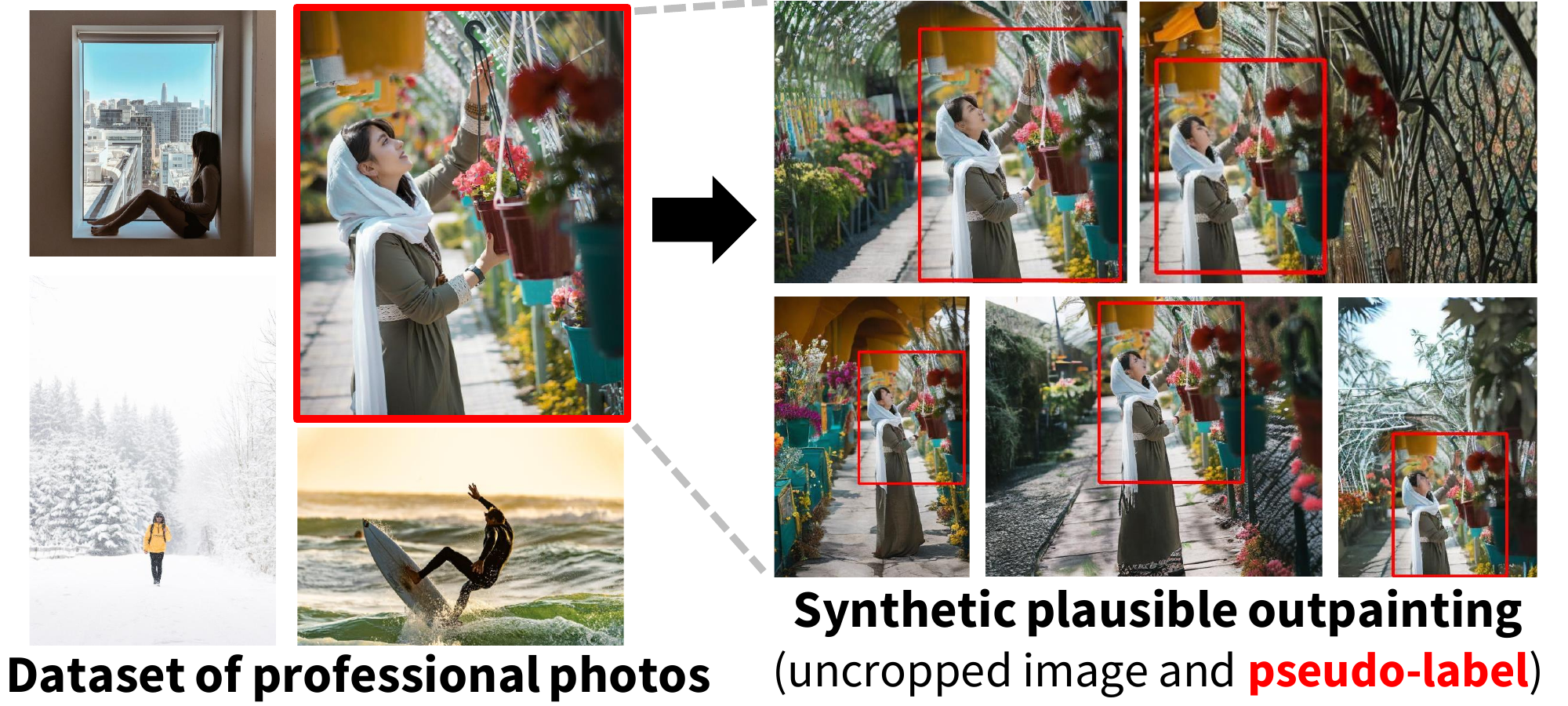}
    \caption{{\bf Generated training pairs.}
    We outpaint professional images (left) to obtain plausible, uncropped input images (right).
    The original image is treated as a pseudo-label crop (red).
    Since the images come from stock image collections, each pseudo-label is an acceptable, professional crop.
    }
    \label{fig:teaser}
\end{figure}

%% file: src/related.tex
\section{Background and Related Work}
\label{sec:related}

Prior works on image cropping have proposed a broad set of methods, which include optimizing for attention/saliency~\cite{cropcomplexity,flms}, heuristics~\cite{templates}, and user-interaction~\cite{gaze}.
Recently, end-to-end data-driven approaches have shown strong performance on benchmarks.
\OURMETHOD follows this paradigm and we compare to that body of work.

\textbf{Data-driven approaches for cropping.}
Most recent works~\cite{cacnet,gaic,hcic,globalview,transview,rankconsistentcrop} utilize {\bf direct supervision} and require large amounts of human-annotated crops and crop scores to train.
GAIC~\cite{gaic}, CPC~\cite{cpc}, and FCDB~\cite{fcdb} are commonly used datasets.
These datasets are expensive to annotate (CPC and GAIC have 259K and 106K crops; on average 24 and 86 crops per image) and the quality of these images and crowd-sourced crops can vary (see \S~\ref{sub:supp_dataset_comparison}).
Since these datasets are not specific to a subject type (e.g., human), prior work on human-centric cropping~\cite{hcic} is limited by the number of images available for training and evaluation.
Only 1.1K, 339, 176, and 39 of the images in CPC, GAICD, FCDB and FLMS~\cite{flms} are human-centric~\cite{hcic}; the human-centric evaluation sets consist of only 50 images from GAICD and 215 from FLMS and FCDB combined.
SACD~\cite{sacd} is a recent dataset for subject-aware cropping that does not focus on a particular subject type but contains 24K+ labels and 5.2 million ranking pairs generated using their annotation procedure.
Our approach differs in that we generate a dataset to provide weak-supervision for a given subject type.
We focus on the subject-aware task because cropping better portraits of people is a subtle and important use case, and choosing a subject type (e.g., people) is a simple way to select the most relevant portion of a stock image dataset for the task.
Extra experiments in \S\ref{sub:quantitative_eval} and \S\ref{sub:qualitative_eval} suggest that subject-type can be a general object category and that \OURMETHOD can generalize to subject types not targeted during training.

\OURMETHOD is comparable to VFN~\cite{vfn}, which is also weakly-supervised by high-quality professional images; VFN generates likely bad crops from within these good images to form ranking pairs.
We compare against VFN trained on our stock images and find that \OURMETHOD performs better, showing that our generated dataset is richer than the ranking pairs mined by VFN.

Other works have also experimented with weak or external supervision in addition to the fully-supervised data.
Despite being human-centric, HCIC~\cite{hcic} trains on all of the images and annotations in CPC and GAICD.
CACNet~\cite{cacnet} utilizes a second composition classification dataset~\cite{kupcp}.
\cite{rankconsistentcrop} trains on the test images, without labels.
\OURMETHOD is only weakly-supervised --- we only train on the uncropped, pseudo-labeled data generated by our pipeline.

\cite{outwardcrop} uses outpainting to enlarge the set of possible crops at test time but utilizes GAICD for training.

Models for data-driven approaches can be categorized by which variant of the cropping task they attempt to solve: (1) learning to rank a set of crop candidates~\cite{hcic,gaic,cpc,transview} and (2) regressing crops directly~\cite{cacnet,globalview}.
\OURMETHOD directly regresses crops and we use an architecture inspired by CACNet~\cite{cacnet}.
Like other regression approaches, we quantitatively evaluate using the standard Intersection-over-Union (IoU) and boundary displacement (Disp) metrics against human-annotated crops.
We do not focus on ranking metrics (e.g., SRCC) but provide those results in \S~\ref{sub:crop_ranking}.

\textbf{Dataset generation using text-image diffusion.}
\cite{fakeitmakeit,stablerep} use Stable Diffusion (SD)~\cite{stablediffusion} to synthesize data for ImageNet~\cite{imagenet} and other generic image classification tasks.
InstructPix2Pix~\cite{instructpix2pix} generates a dataset for text-driven image edits.
\OURMETHOD also leverages the image generation capabilities of SD but to transform stock images into labels for cropping.
Techniques to generate data for other spatial tasks (e.g., object grounding or placement) are interesting future work.

Other works have studied the {\bf outpainting task} directly~\cite{ganextend,verylongscenery,outpaintganinversion}.
Recently, outpainting using text-image diffusion~\cite{dalle2,firefly} has been shown to be a powerful interactive tool in the hands of artists.
Future work on automatic outpainting would benefit the quality of the training data produced by our pipeline.

\textbf{Pre-trained models used by~\OURMETHOD.}
In addition to Stable Diffusion (SD)~\cite{stablediffusion}, we use other off-the-shelf models in our pipeline.
These include an image captioner~\cite{blip2} and an instance segmenter~\cite{yolov8}.
Text-conditioning, even with noisy captions, helps SD produce more plausible outpainted images.
The instance segmenter is used to detect and segment the subject as an input to our model. While the YOLOv8 model that we use is limited to the COCO~\cite{coco} object classes, we anticipate that advances in models such as SAM~\cite{segmentanythingmodel} will enable arbitrary subject classes.

%% file: src/method.tex
\input{src/figure/pipeline}
\input{src/figure/outpaint_bad}
\input{src/figure/architecture}

\section{Methods}
\label{sec:methods}

Given an image, \OURMETHOD generates possible crop rectangles that lead to aesthetically pleasing compositions.
Our method is ``subject-aware''. We condition the cropper on both the input image and an estimated pixel mask denoting the location of a given subject.
The core of our approach is to generate a dataset of cropped and uncropped image training pairs using a pre-trained generative model (\S\ref{sub:dataset_generation}).
We then use the generated data to train a subject-aware cropping model  (\S\ref{sub:cropping_architecture}).

\subsection{Dataset Generation with Image Outpainting}
\label{sub:dataset_generation}

Our first goal is to construct a dataset of image pairs, one casually-framed and one expertly-framed, to supervise our cropping model.
Because our dataset generation is automated, we synthesize images according to a subject type (i.e., cropping portraits like professional portraits).
We filter the stock image collection~\cite{unsplash} to a set of relevant images.
These photos, by nature of their inclusion in the stock photo collection, have been vetted for good composition and high aesthetic quality by human experts.
We then use a pre-trained diffusion model to hallucinate plausible out-of-frame content for each image.
Fig.~\ref{fig:synthesis_pipeline} illustrates our pipeline and a generated cropped-uncropped training pair.

For each stock image, we apply the following operations:
\begin{enumerate}
    \item \textit{Pre-processing and filtering.}
    We filter for images that include an identifiable subject (e.g., person in portraiture; Fig.~\ref{fig:synthesis_pipeline}a).
    This is done with metadata tags first and then with an object detector~\cite{yolov8}.
    We also discard the image if it contains too many possible subjects (e.g., $>5$).
    For simplicity, if there are multiple possible subjects, we select the largest one as the dominant subject (by bounding box area).
    \item \textit{Estimating image captions.}
    We use an automatic image captioning algorithm~\cite{blip2} to estimate a text-conditioning string: $s$ (Fig.~\ref{fig:synthesis_pipeline}b).
    The purpose of this text conditioning is to constrain the content generated by the diffusion model.
    Omitting it can lead to unrelated contexts in the outpainting; see Fig.~\ref{fig:supp_no_blip}.

    \item \textit{Outpainting.}
    We randomly downscale the image with bilinear interpolation and paste it into a surrounding 512$\times$512 canvas to obtain an image $\mathbf{x}$ (Fig.~\ref{fig:synthesis_pipeline}c).
    We also compute a binary mask $\mathbf{m}$ with 1's in the area corresponding to valid pixels.
    We then pass $\mathbf{x}$ and $\mathbf{m}$ to a pre-trained diffusion inpainting model~\cite{stablediffusion} to obtain a 512$\times$512 outpainted image (Fig.~\ref{fig:synthesis_pipeline}d):
    \[
        \mathbf{x'} := \texttt{StableDiffusionInpaint}(\mathbf{x}, \mathbf{m}, s)
    \]
    Because we wish to learn to crop images of different aspect ratios, not just the 1:1 square images produced by Stable Diffusion,
    we sample a rectangular crop $\mathbf{x}_o$ from inside $\mathbf{x}'$ that also encloses the original image.
    $\mathbf{x}_o$ is chosen to have a common aspect ratio (e.g., 2:3, 4:5, etc.). We refer to the coordinates of the original image region in $\mathbf{x}_o$ as $\mathbf{y}\in \mathbb{R}^4$ and treat this as a crop pseudo-label.
    We also run an object segmenter~\cite{yolov8} to update the subject's bounding box (since a cropped subject may grow in size due to outpainting) and to produce the subject mask, $\mathbf{m}_o$, needed for \OURMETHOD.
    Together, $(\mathbf{x}_o, \mathbf{m}_o)$ and $\mathbf{y}$ form a weakly-labeled training pair.
    (Fig.~\ref{fig:synthesis_pipeline}f).

    \item \textit{Outpainting quality filtering.}
    Not all outpainting attempts lead to plausible images (see Fig.~\ref{fig:bad_outpaint});
    we discard the most striking failure cases (Fig.~\ref{fig:synthesis_pipeline}e) using two automatic heuristics described later in this section.
\end{enumerate}
\noindent We repeat the pipeline multiple times per image to sample additional variations for data amplification (4$\times$ to 8$\times$ depending on the initial number of stock images available).

\subsubsection{Filtering the uncropped images.}

We discard two prominent classes of images where Stable Diffusion (SD)~\cite{stablediffusion} fails to outpaint a reasonable scene.
\begin{enumerate}
    \item Images with a new subject (e.g., another person) in the outpainted region (Fig.~\ref{fig:bad_outpaint}a).
    This can happen if the subject or if the original region (in $\mathbf{x}_o$) is small.
    \item Images depicting a grid of images or scenes that ``frame'' the original image (Fig.~\ref{fig:bad_outpaint}b).
    These images leak information about the crop label.
\end{enumerate}
These images account for about 20\% of the outpainted portrait images.
To suppress images of category (1), we run a heuristic that compares the size of the largest object of the subject class (e.g., person) in the outpainted region with the size of the originally identified subject.
If another object has an area larger than $\frac{1}{4}$ of the subject's area, the image is discarded.
To address category (2), we separately train a binary classifier, $D_{quality}$, to identify grid, composite, and bordered images.
$D_{quality}$ is a standard ResNet-50~\cite{resnet} that we trained on 3K generated images (of which 15\% are `bad') and it identifies approximately 90\% of the images with such issues (details in \S~\ref{sub:supp_dataset_generation_impl}).
\OURMETHOD can still train without these filters, but we show via ablations in \S~\ref{sub:data_filtering} that removing these problematic data boosts results slightly (up to 0.03 IoU and 0.008 Disp).

Images produced by current text-image diffusion models often have artifacts (e.g., in fine details such as faces; zoom into Fig.~\ref{fig:teaser}-right).
With Stable Diffusion inpainting, these artifacts appear even in the non-inpainted regions; we do not correct them and find that they do not impede training of the cropping model.
Blending the original (cropped) image into the outpainted synthetic image would reduce artifacts, but it would also leak information about the crop pseudo-label due to the absence of artifacts~\cite{spotcnngenerated}.


\subsection{Cropping Model and Training}
\label{sub:cropping_architecture}

We describe the cropping model used in our experiments.

\subsubsection{Model inputs and outputs.}
The input is an uncropped image $\textbf{x}_o$ and subject mask $\textbf{m}_o$, which contain the box specified by the crop pseudo-label $\mathbf{y}$.
We scale and zero pad $\textbf{x}_o$ and $\textbf{m}_o$ to $256\times256$, since our model is not fully convolutional.
The output is a vector $\mathbf{\hat{y}}\in \mathbb{R}^4$.

\subsubsection{Architecture.}

At a high level, our model design is inspired by CACNet~\cite{cacnet}, sharing a similar three-part structure of a multi-scale CNN feature extractor and two branches: one for regressing crops at a grid of anchor points (`cropping branch') and the other for producing blending weights for the crops (`composition branch').

The feature extractor is a ResNet-50 that accepts four-channel inputs: the RGB image and the binary subject mask.
From the 256$\times$256 inputs, we obtain a 16$\times$16 grid of features.
Our cropping branch is a small 2-layer transformer-encoder~\cite{transformer} that can
easily learn global interactions between distant parts of the image.
Each of the 16$\times$16 transformer inputs and outputs corresponds to an anchor point, and each anchor point produces one crop proposal: $\mathbf{\hat{y}}_{ij}\in \mathbb{R}^4$ at the $ij$'th anchor point.
In our additional experiments (\S\ref{sub:conditional_cropper}), we extend this component to inject conditional control into our cropping model by replacing the transformer-encoder with a transformer-decoder~\cite{transformer}.
In our composition branch, we predict blending weights for the crop proposals.
Since we want to prioritize the subject, we use zero weights for proposals by anchor points that lie outside the subject's bounding box.
For anchor points within the subject bounding box, we use RoIAlign~\cite{maskrcnn} and RoDAlign~\cite{gaic} layers to pool the spatial CNN features from inside and outside of a crop proposal, similar to prior cropping work~\cite{hcic}.
A feed-forward network uses these features to produce a weight: $w_{ij}$ for the $ij$'th proposal.
The final crop prediction, $\mathbf{\hat{y}}$, is the softmax-weighted sum:
\[
    \mathbf{\hat{y}} = \sum_{i=1}^{16}\sum_{j=1}^{16} \texttt{Softmax}(\mathbf{w})_{ij}\mathbf{\hat{y}}_{ij}
\]
Fig.~\ref{fig:architecture} shows our architecture and \S~\ref{sub:supp_model_and_training} provides implementation details and comparison to CACNet.

\subsubsection{Losses.}

We train using a \textit{regression loss}: the L1 error between $\mathbf{\hat{y}}$ and $\mathbf{y}$.
We also add two additional losses.
\begin{enumerate}
    \item \textit{Per-anchor regression loss.}
    We apply L1 loss between the predicted crop at each anchor point and $\mathbf{y}$, with a low loss weight ($\frac{1}{10}$).
    This reduces variance by encouraging all anchors to make predictions similar to the ground truth (i.e., reasonably shaped boxes).

    \item \textit{Subject boundary loss.}
    A common error that we notice when training on the aforementioned losses is that
    the model produces crops that cut subjects at their extremities (e.g., tip of the feet; Fig.~\ref{fig:bad_crop_examples}-left).
    This is unflattering and is difficult to penalize because the predicted crop can be within pixels of the true crop.
    Manually-annotated ranking datasets may explicitly label such crops as bad~\cite{cpc}.
    Without relying on manual annotations, we introduce a margin L1 loss to discourage crops within 2.5\% of the subject mask's bounding box.
    To avoid overriding real labels, this loss is applied only if the label also does not crop on or near the subject.
\end{enumerate}

\subsubsection{Implementation.}
We optimize the network end-to-end using AdamW \cite{adam} and cosine annealing~\cite{cosinelr}.
In addition to the input processing described above, we apply standard image augmentations (e.g., flip, color jitter, blur, distortion, etc.).
See \S~\ref{sub:supp_model_and_training} for hyper-parameters and details.

%% file: src/figure/pipeline.tex
\begin{figure*}[t]
    \centering
    \includegraphics[width=\textwidth]{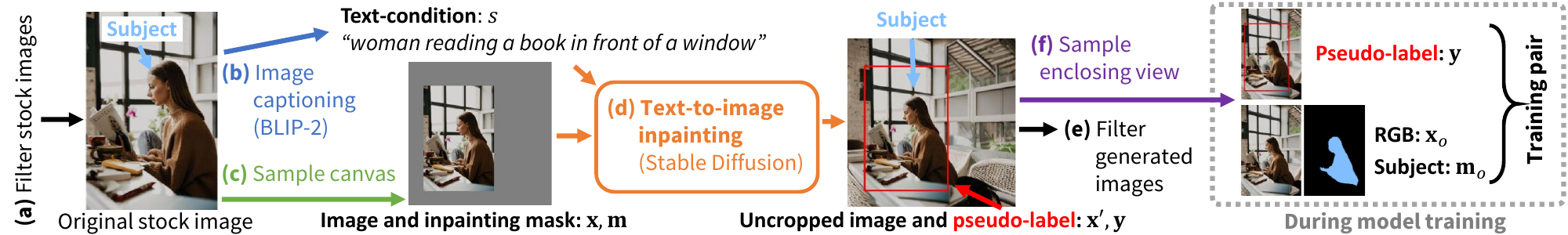}
    \caption{{\bf Dataset generation pipeline.}
    Stages are marked (a-f).
    Refer to \S\ref{sub:dataset_generation} for detailed explanation.
    We start with a stock image (a) and estimate its text caption (b).
    To determine the region to be outpainted, we sample a blank canvas to outpaint around the image (c).
    Outpainting is done using a text-to-image inpainting model~\cite{stablediffusion} and results in a square image (d).
    Afterwards, we apply automated filters to remove poorly generated images (e).
    Later, when training a cropping model, we sample an enclosing view (f) in the uncropped image from a common aspect (e.g., $3:4$) so that the model generalizes beyond square images.
    The region containing the original image is treated as a pseudo-label when training a cropping model.
    }
    \label{fig:synthesis_pipeline}
\end{figure*}

%% file: src/figure/outpaint_bad.tex
\begin{figure}[t]
    \centering
    \begin{subfigure}[t]{\columnwidth}
        \centering
        \includegraphics[width=0.24\columnwidth]{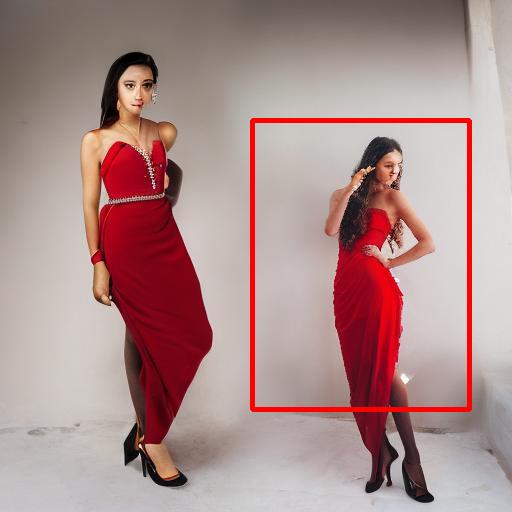}
        \includegraphics[width=0.24\columnwidth]{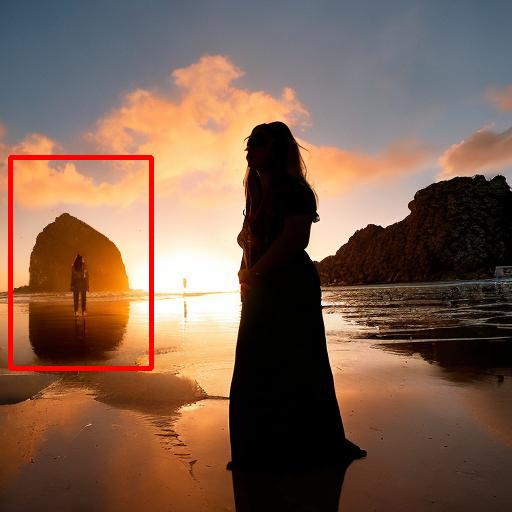}
        \includegraphics[width=0.24\columnwidth]{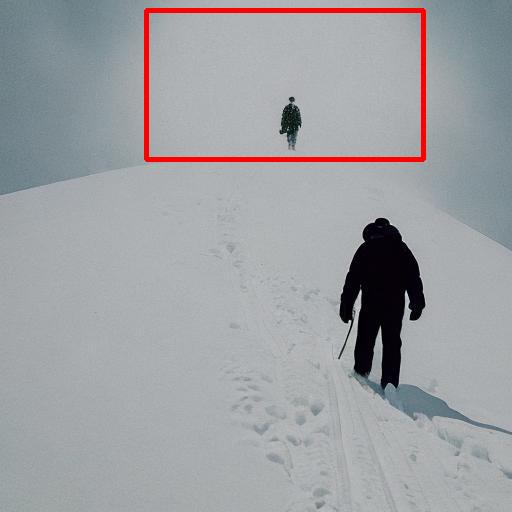}
        \includegraphics[width=0.24\columnwidth]{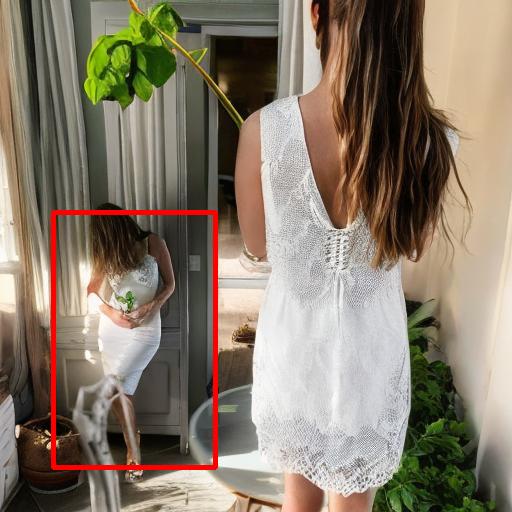}
        \caption{Additional outpainted subject}
    \end{subfigure}

    \vspace{0.2em}

    \begin{subfigure}[t]{\columnwidth}
        \centering
        \includegraphics[width=0.24\columnwidth]{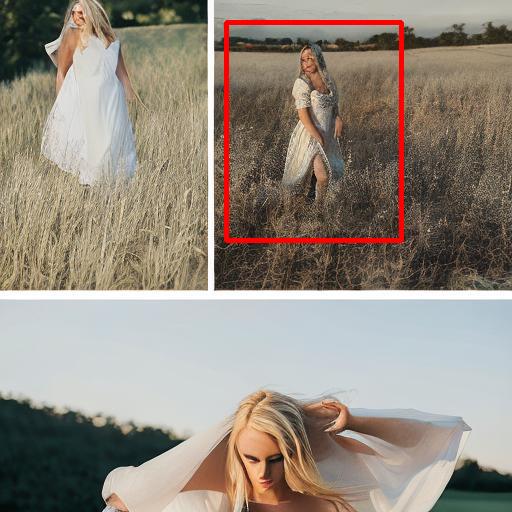}
        \includegraphics[width=0.24\columnwidth]{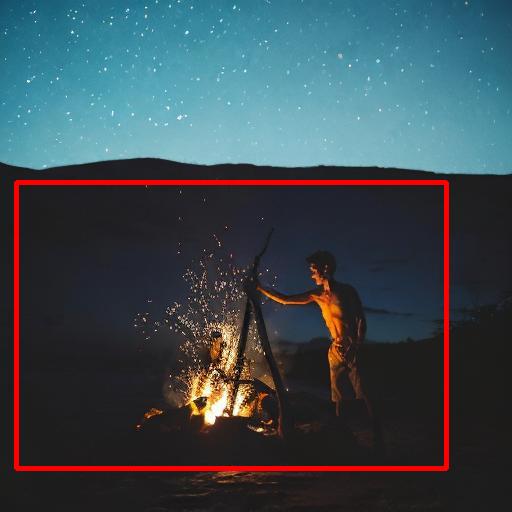}
        \includegraphics[width=0.24\columnwidth]{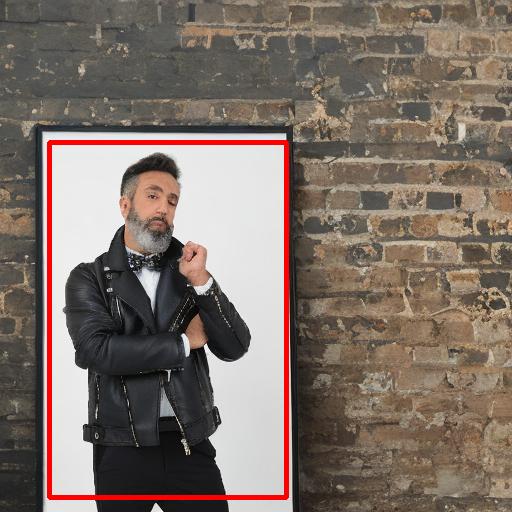}
        \includegraphics[width=0.24\columnwidth]{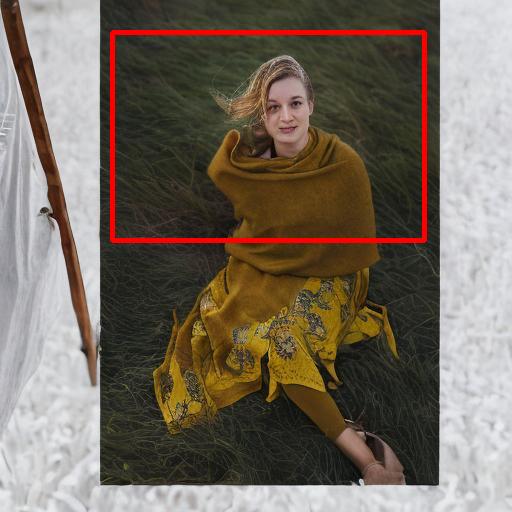}
        \caption{Tiled, composite, or framed images}
    \end{subfigure}
    \caption{{\bf Common outpainting failure cases.} The original image is marked with a red rectangle.
    (a) An extra person was synthesized in the outpainted region. This can alter the ideal composition of the scene.
    (b) The outpainted region is a grid or composite of multiple images (col 1, 2), frames the original image (col 3), or has a border (col 4). These artificial edges can bias the model towards detecting sharp borders.}
    \label{fig:bad_outpaint}
\end{figure}

%% file: src/figure/architecture.tex
\begin{figure*}[t]
    \centering
    \includegraphics[width=\textwidth]{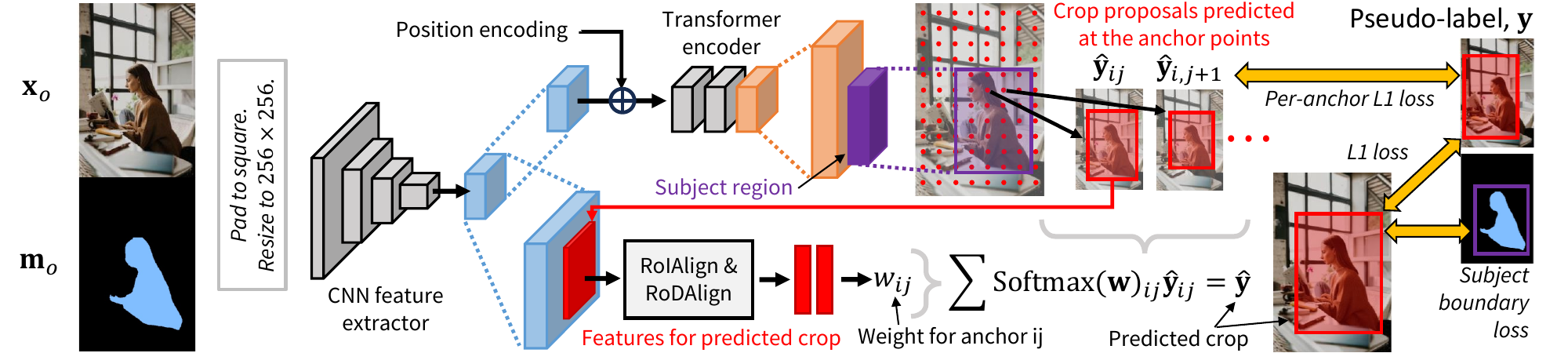}
    \caption{{\bf Cropping model architecture.}
    Our design is inspired by CACNet~\cite{cacnet}: details in~\S\ref{sub:cropping_architecture} and \S~\ref{sub:supp_model_and_training}.
    We extract CNN features from the input image, $\mathbf{x}_o$, and subject mask, $\mathbf{m}_o$. These features are used by a transformer-encoder~\cite{transformer} to generate crop proposals at a grid of anchor points.
    The crop proposal at each anchor point contained in the subject region is weighted by a second branch.
    A softmax-weighted sum computes the final crop prediction $\mathbf{\hat{y}}$.
    }
    \label{fig:architecture}
\end{figure*}

%% file: src/dataset.tex
\section{Stock Image Dataset}
\label{sec:dataset}

\input{src/table/dataset.tex}

Unsplash contains over three million curated images and is publicly accessible for research use~\cite{unsplash}.
While our primary motivation is to crop human portraits, we also experiment with five other categories: cats, dogs, horses, birds, and cars.
We select the relevant images using provided metadata and off-the-shelf object detection as described in~\S\ref{sub:dataset_generation}.
Because Unsplash reflects a real-world distribution of images and subject matter submitted by photographers, the amount of data by subject varies.
After filtering, we are left with 73K, 8K, 11K, 2.8K, 12K and 11K images for portraits, cats, dogs, birds, horses, and cars.
We designate a fraction of the images for test and validation; see Tab.~\ref{tab:dataset}.

\subsection{Evaluation Sets for Subject-Aware Cropping}
\label{sub:new_eval_sets}

Prior cropping datasets such as FLMS~\cite{flms}, FCDB~\cite{fcdb}, and SACD~\cite{sacd} lack the quantity of images in any particular subject category needed to serve as evaluation (having only 500, 348, and 290 test images total).
In order to evaluate \OURMETHOD, we construct new evaluation sets for the six aforementioned subjects, derived from the Unsplash testing images.
We select Unsplash images with space for tighter crops (alternative framings) and task the model to produce crops that preserve the aesthetic qualities of a good composition;
{\em a good cropping model should not produce bad crops that violate compositional norms} (e.g., by cutting through a person at a joint).
Therefore, to create an evaluation set, one author of this paper, who is a photography domain expert, annotated crops for 1,905 images of the six aforementioned subjects (see Tab.~\ref{tab:dataset}), producing 2.3 good crops per image on average.
The annotations for this data and the images are all publicly available.
We also use these images for qualitative evaluation in \S\ref{sub:qualitative_eval}, where we measure the rate at which cropping methods produce common framing errors.

%% file: src/table/dataset.tex
\begin{table}[t]
    {\centering \small \setlength{\tabcolsep}{5pt}
    \begin{tabularx}{\columnwidth}{lrr|r|rr}
        \toprule
        & \multicolumn{2}{c}{Train} & \multicolumn{1}{c}{Val} & \multicolumn{2}{c}{Test} \\
        \multicolumn{1}{c}{Subject} & \multicolumn{1}{c}{\# imgs} & \multicolumn{1}{c}{\# outpainted} & \multicolumn{1}{c}{\# imgs} & \multicolumn{1}{c}{\# imgs} & \multicolumn{1}{c}{\# labeled} \\
        \midrule
        Human    & 52.0K & 188K ($3.6\times$) & 11.0K & 10.3K & 1000 \\
        Cat      & 6.7K & 18K ($2.7\times$) & 838 & 804 & 204 \\
        Dog      & 8.9K & 24K ($2.7\times$) & 1.1K & 1.2K & 200 \\
        Bird     & 9.0K & 25K ($2.8\times$) & 1.3K & 1.8K & 201 \\
        Horse    & 2.2K & 12K ($5.3\times$) & 221 & 412 & 200 \\
        Car      & 9.2K & 23K ($2.5\times$) & 891 & 811 & 100 \\
        \bottomrule
    \end{tabularx}
    }
    \caption{{\bf Dataset statistics and splits for each subject class.} \# outpainted is the number of synthetic training images that pass the automatic quality filters in \S\ref{sub:dataset_generation}. \# labeled is the size of our hand-labeled evaluation subset (of the test split).}
    \label{tab:dataset}
\end{table}

%% file: src/result.tex
\input{src/table/existing_iou_disp.tex}
\input{src/table/our_iou_disp.tex}
\input{src/table/violation}

\section{Experiments}
\label{sec:results}

In \S\ref{sub:quantitative_eval}, we evaluate \OURMETHOD quantitatively on the existing subject-aware cropping benchmarks~\cite{sacd,hcic} and our class-specific test sets described in \S\ref{sub:new_eval_sets}.
In order to control subjectivity and to provide clearer insight into cropping model failures, we measure the violation rate for crops on a set of five pre-determined, objective aesthetic-quality guidelines (\S\ref{sub:qualitative_eval}).
We also conduct additional experiments and ablations in \S\ref{sub:conditional_cropper} and \S\ref{sub:ablation_and_additional}.

\subsection{Quantitative Evaluation}
\label{sub:quantitative_eval}

\subsubsection{Metrics.}
Intersection-over-Union (IoU) and boundary displacement (Disp) are common metrics for cropping evaluation used in prior work~\cite{hcic,sacd}.
If there are multiple ground-truth labels in an image, we follow the standard protocol of using the label with the top IoU to evaluate the prediction.
While IoU and displacement metrics are not fully informative of cropping quality, they provide a standardized way to compare to existing methods.

\subsubsection{Baselines.}
We compare \OURMETHOD to HCIC~\cite{hcic}, CACNet~\cite{cacnet}, and VFN~\cite{vfn} and reported results in prior work~\cite{sacd}.

HCIC and CACNet are recent, supervised methods with publicly available code;
HCIC is trained on CPC~\cite{cpc} or GAICD~\cite{gaic} and is subject-aware, while CACNet is trained on FCDB~\cite{fcdb} and KUPCP~\cite{kupcp}.
VFN~\cite{vfn} is weakly supervised (though it can also be trained in a supervised manner on CPC).
For direct comparison to \OURMETHOD, we train VFN on Unsplash images, including our subject masks and using the same ResNet-50 backbone.

\input{src/figure/bad_crop_examples}
\input{src/figure/condition}

\subsubsection{Comparison on prior benchmarks.}
Tab.~\ref{tab:result_existing} compares our \OURMETHOD to prior work on existing, published benchmarks.
FCDB + FLMS is the human-centric test set used by~\cite{hcic}.
SACD is the test set from~\cite{sacd}, which also contains non-human images.
We use \OURMETHOD trained on outpainted human portraits for these experiments.

\OURMETHOD is significantly better than the comparable VFN (up to 0.15 IoU and 0.04 Disp) and is competitive with supervised methods on both datasets.
\OURMETHOD is within 0.014 IoU and 0.004 Disp to HCIC on FCDB + FLMS and better than HCIC, GAIC, and CACNet on SACD (by around 0.02 IoU and 0.005 Disp).
On SACD, \OURMETHOD is within 0.036 IoU and 0.014 Disp of SAC-Net~\cite{sacd}, a method tailored around the human-annotated label structure in the SACD training data.
Despite SACD not being human-subject exclusive, \OURMETHOD trained on human portraits is able to show generalization ability --- more so than other supervised methods trained on GAICD, CPC, and FCDB.

\subsubsection{Comparison on the Unsplash test sets from \S\ref{sub:new_eval_sets}.}
Tab.~\ref{tab:result_ours} shows results on the six categories: humans, cats, dogs, birds, horses, and cars.
\OURMETHOD is trained on generated data filtered by subject category, and it
significantly outperforms VFN, while remaining competitive with supervised methods.
We also test two additional versions of \OURMETHOD: (1) trained jointly on the six categories (\OURMETHOD-6) and (2) trained on humans only (\OURMETHOD-H).
All three variations of \OURMETHOD perform similarly, showing that specialization of the synthetic dataset to the subject category is not necessary (though it can obtain similar results with less data).

\subsection{Qualitative Evaluation}
\label{sub:qualitative_eval}

Image cropping is challenging to evaluate as the best crop is subjective. Quantitative metrics such as IoU and Disp are also not fully informative~\cite{gaic}.
Narrowing the evaluation by subject type (e.g., human portraits) can be more objective since there are well-established technical rules on what makes a bad crop.
For instance, cropping a person through a joint (e.g., ankle, knee, elbow) is generally regarded as unflattering~\cite{popphotobook,stunningbook}.
We perform a qualitative evaluation that counts the number of such violations in a sample of 100 images per class.
Specifically, we consider five (non-mutually-exclusive) questions: {\it Does the crop:
\begin{enumerate}
    \item cut unnaturally through the subject?
    \item cut unnaturally through the scene (e.g., other objects)?
    \item have too much or too little negative space?
    \item have or create unnecessary clutter around the edges?
    \item lack balance (e.g., missing symmetries, rule-of-thirds)?
\end{enumerate}}
\noindent Fig.~\ref{fig:bad_crop_examples} shows two crops with subtle mistakes (to casual observers).
Please refer to \S~\ref{sub:full_qualitative} for more detailed explanations and additional visual examples.

While these five criteria are not exhaustive, they reflect common errors that we observed and critiques that a poorly-composed image might receive.
Also, although experts may deliberately violate these rules for artistic expression, we observe empirically that cropping model failures on these criteria are indicative of bad crops.

Tab.~\ref{tab:qualitative_results} and Tab.~\ref{tab:qualitative_results_by_type} compare the methods by the subject category and the violation type, respectively.
Like in Tab.~\ref{tab:result_ours}, we compare the default (specialized) \OURMETHOD, \OURMETHOD-H trained on human images only, and \OURMETHOD-6 trained jointly on all six classes.
All three versions of \OURMETHOD are better than or competitive with (supervised) HCIC and CACNet.
\OURMETHOD-6 is the best in five of six subjects and overall, suggesting that more training data is helpful.
This can be seen on the horse class (which has the fewest training images), where the specialized \OURMETHOD makes 2$\times$ as many errors as HCIC, but \OURMETHOD-6 makes only 0.6$\times$ as many.

By violation type, all three versions of \OURMETHOD do well at preserving the subject (V1) and managing negative space (V3). However, HCIC and CACNet are better at preserving context (V2) and managing edge clutter (V4).

\subsection{Extension: Conditioning Signal}
\label{sub:conditional_cropper}

A common limitation of methods that directly regress a crop is that they lack user control: one cannot specify an aspect ratio or tightness.
We propose a small extension of \OURMETHOD that is conditioned on the above properties and hypothesize that, given a large number of training images, \OURMETHOD-C can disentangle aspect ratio and tightness (area) while still learning to predict good crops.
To achieve this, we swap the transformer-encoder in \OURMETHOD with a transformer-decoder~\cite{transformer} and encode the conditioning with a feed-forward network (details in \S~\ref{sub:supp_model_variants}).
Fig.~\ref{fig:conditional_cropper} shows examples with different conditioning applied.
While \OURMETHOD-C does not enforce exact adherence to the signal, it does generally respect orientation (portrait vs. landscape), and changing the signal varies the crops (in aspect and tightness).
We anticipate that future work may improve adherence or could learn to condition on other interesting properties from the data (e.g., composition patterns).

\subsection{Additional Results, Ablations \& Images}
\label{sub:ablation_and_additional}
Please refer to \S~\ref{sub:throughput_cost} -- \S~\ref{sub:crop_ranking} for additional analysis (e.g., computational cost, different cropping model architectures, results on cropping generic images, ranking metrics), ablations (e.g., subject-awareness, our data-quality filters, training dataset size), and example crops.

%% file: src/table/existing_iou_disp.tex
\begin{table}[t]
    {\centering \small \setlength{\tabcolsep}{2.4pt}
    \begin{tabularx}{\columnwidth}{lccrrrr}
        \toprule
        \multicolumn{3}{c}{}
            & \multicolumn{2}{c}{FCDB+FLMS}
            & \multicolumn{2}{c}{SACD} \\
        \multicolumn{1}{c}{Method} & \multicolumn{1}{c}{Trained on}
            & \multicolumn{1}{c}{WS}
            & \multicolumn{1}{c}{IoU $\uparrow$} & \multicolumn{1}{c}{Disp $\downarrow$}
            & \multicolumn{1}{c}{IoU $\uparrow$} & \multicolumn{1}{c}{Disp $\downarrow$} \\
        \midrule
        VFN & CPC &
            & \fromhcic0.6509 & \fromhcic0.0876
            & - & - \\
        LVRN & CPC &
            & \fromhcic0.7373 & \fromhcic0.0674
            & \fromsacd0.6962 & \fromsacd0.0765 \\
        GAIC & CPC &
            & \fromhcic0.7260 & \fromhcic0.0708
            & - & - \\
        GAIC & GAICD &
            & - & -
            & \fromsacd0.7124 & \fromsacd0.0696 \\
        CGS & CPC &
            & \fromhcic0.7331 & \fromhcic0.0689 & - & - \\
        \multicolumn{3}{l}{CACNet\hspace{0.1in}FCDB,KUPCP}
            & \fromhcic0.7364 & \fromhcic0.0676
            & 0.7109 & 0.0716 \\
        HCIC & GAICD &
            &- & -
            & 0.7120 & 0.0683 \\
        HCIC & CPC &
            &\fromhcic\best{0.7469} & \fromhcic\best{0.0648}
            & 0.7109 & 0.0712 \\
        FRCNN-m & SACD & & - & -
            & \fromsacd0.7306 & \fromsacd0.0587 \\
        SAC-Net & SACD & & - & -
            & \fromsacd\best{0.7665} & \fromsacd\best{0.0491} \\
        \midrule
        VFN & Flickr & \checkmark
            & \fromhcic0.5115 & \fromhcic0.1257
            & \fromsacd0.6690 & \fromsacd0.0887 \\
        VFN & Unsplash & \checkmark &
            0.5783 & 0.1114 & 0.6555 & 0.0775 \\
        \OURMETHOD & Unsplash & \checkmark &
            \best{0.7334} & \best{0.0687}
            & \best{0.7301} & \best{0.0632} \\
        \bottomrule
    \end{tabularx}
    }
    \caption{{\bf Quantitative comparison on existing benchmarks.}
        The best result in each category is \best{bold}; WS indicates weakly-supervised.
        \fromhcic~and \fromsacd~are results reported by~\cite{hcic} and~\cite{sacd}, respectively.
        FCDB+FLMS is the human-centric test-set used by~\cite{hcic}.
        SACD is the subject-aware dataset from~\cite{sacd}, which includes non-human subjects.
        \OURMETHOD refers to our model trained on outpainted human portraits.
    }
    \label{tab:result_existing}
\end{table}

%% file: src/table/our_iou_disp.tex
\begin{table*}[t]
    {\centering \small \setlength{\tabcolsep}{3.82pt}
    \begin{tabularx}{\textwidth}{lcccccccccccccc}
        \toprule
        \multicolumn{3}{c}{}
            & \multicolumn{2}{c}{Human}
            & \multicolumn{2}{c}{Cat}
            & \multicolumn{2}{c}{Dog}
            & \multicolumn{2}{c}{Bird}
            & \multicolumn{2}{c}{Horse}
            & \multicolumn{2}{c}{Car} \\
        \multicolumn{1}{c}{Method} & Trained on & Weak sup
            & \multicolumn{1}{c}{IoU $\uparrow$} & \multicolumn{1}{c}{Disp $\downarrow$}
            & \multicolumn{1}{c}{IoU $\uparrow$} & \multicolumn{1}{c}{Disp $\downarrow$}
            & \multicolumn{1}{c}{IoU $\uparrow$} & \multicolumn{1}{c}{Disp $\downarrow$}
            & \multicolumn{1}{c}{IoU $\uparrow$} & \multicolumn{1}{c}{Disp $\downarrow$}
            & \multicolumn{1}{c}{IoU $\uparrow$} & \multicolumn{1}{c}{Disp $\downarrow$}
            & \multicolumn{1}{c}{IoU $\uparrow$} & \multicolumn{1}{c}{Disp $\downarrow$} \\
        \midrule
        CACNet & FCDB,KUPCP & &
            0.749 & 0.062 &
            0.740 & 0.065 &
            0.742 & 0.063 &
            0.696 & 0.076 &
            0.757 & 0.060 &
            0.727 & 0.068 \\
        HCIC & GAICD & &
            0.723 & 0.065 &
            0.733 & 0.065 &
            0.740 & 0.061 &
            0.696 & 0.074 &
            0.754 & \best{0.059} &
            0.730 & \best{0.064} \\
        HCIC & CPC & &
            \best{0.750} & \best{0.060} &
            \best{0.769} & \best{0.056} &
            \best{0.759} & \best{0.057} &
            \best{0.714} & \best{0.069} &
            \best{0.759} & \best{0.059} &
            \best{0.735} & 0.065 \\
        \midrule
        VFN & Unsplash & \checkmark &
            0.622 & 0.095 &
            0.633 & 0.093 &
            0.621 & 0.095 &
            0.573 & 0.106 &
            0.623 & 0.093 &
            0.633 & 0.088 \\
        \OURMETHOD & Unsplash & \checkmark &
            \best{0.750} & \best{0.061} &
            \best{0.777} & \best{0.054} &
            \best{0.758} & \best{0.058} &
            \best{0.712} & \best{0.071} &
            \best{0.757} & \best{0.059} &
            \best{0.744} & \best{0.063} \\
        \midrule
        \multicolumn{2}{l}{\OURMETHOD-6 (all 6)} & \checkmark &
            \best{0.752} & \best{0.060} &
            0.767 & 0.057 &
            0.748 & 0.061 &
            \best{0.719} & \best{0.069} &
            \best{0.760} & \best{0.058} &
            0.742 & 0.062 \\
        \multicolumn{2}{l}{\OURMETHOD-H (humans)} & \checkmark &
            0.750 & 0.061 &
            0.767 & 0.056 &
            0.751 & 0.059 &
            0.711 & 0.070 &
            0.748 & 0.061 &
            \best{0.748} & \best{0.060} \\
        \bottomrule
    \end{tabularx}
    }
    \caption{{\bf Quantitative comparison on different subject types.}
    Best results per category are \best{bold}.
    We evaluate HCIC~\cite{hcic} without its human-specific feature partitioning scheme for non-human subjects.
    \OURMETHOD, trained on synthesized data of the subject category (middle), is competitive with supervised methods (top).
    To test whether specialization of the training data is necessary, we test \OURMETHOD-6, trained jointly on all six categories, and \OURMETHOD-H, trained on humans only but applied to other categories (bottom).
    The results are similar between all three \OURMETHOD variations, suggesting a degree of generalization.
    }
    \label{tab:result_ours}
\end{table*}


%% file: src/table/violation.tex
\begin{table}[t]
    {\centering \small \setlength{\tabcolsep}{3pt}
    \begin{tabularx}{\columnwidth}{lrrrrrr|r}
        \toprule
        \multicolumn{1}{c}{Method}
            & Human
            & Cat
            & Dog
            & Bird
            & Horse
            & \multicolumn{1}{c}{Car}
            & \multicolumn{1}{c}{Mean} \\
        \midrule
        VFN &
            52 & 41 & 59 & 60 & 55 & 40 & 51.2 \\
        CACNet &
            10 & 9 & 13 & 17 & 23 & 9 & 13.5 \\
        HCIC (CPC) &
            9 & 10 & 12 & 12 & 7 & \best{3} & 8.8 \\
        \OURMETHOD &
            11 & 7 & 5 & 3 & 15 & \best{3} & \best{7.3} \\
        \midrule
        \OURMETHOD-H (human)\hspace{-1em} & 11 & 6 & 5 & 9 & 10 & 4 & 7.5 \\
        \OURMETHOD-6 (all 6) & \best{8} & \best{3} & \best{3} & \best{1} & \best{4} & 8 & \best{4.5} \\
        \bottomrule
    \end{tabularx}
    }
    \caption{{\bf Qualitative results.} Percentage of images with \emph{one or more} violations ($\downarrow$ is better; definitions in \S\ref{sub:qualitative_eval}).
    \OURMETHOD-H and \OURMETHOD-6 are trained on human and all 6 classes.
    See Tab.~\ref{tab:supp_qualitative} for the full breakdown.
    }
    \label{tab:qualitative_results}
\end{table}

\begin{table}[t]
    {\centering \small \setlength{\tabcolsep}{8pt}
    \begin{tabularx}{\columnwidth}{lrrrrr}
        \toprule
        \multicolumn{1}{c}{Method}
            & \multicolumn{1}{c}{V1}
            & \multicolumn{1}{c}{V2}
            & \multicolumn{1}{c}{V3}
            & \multicolumn{1}{c}{V4}
            & \multicolumn{1}{c}{V5} \\
        \midrule
        VFN & 19.7 & 10.1 & 11.0 & 2.6 & 5.3 \\
        CACNet & 6.8 & 1.8 & 4.5 & 1.0 & 1.8 \\
        HCIC (CPC) & 3.7 & \best{0.8} & 2.8 & \best{0.6} & 2.1 \\
        \OURMETHOD & 2.5 & 2.8 & 1.0 & 1.5 & \best{0.6} \\
        \midrule
        \OURMETHOD-H (human)
            & 1.7 & 2.5 & 1.5 & 2.0 & 1.7 \\
        \OURMETHOD-6 (all 6)
            & \best{0.8} & 1.3 & \best{0.3} & 1.7 & 1.0 \\
        \bottomrule
    \end{tabularx}
    }
    \caption{{\bf Qualitative results.} Percentage of images with violations, by violation type and aggregated across the six subject classes ($\downarrow$ is better; definitions in \S\ref{sub:qualitative_eval}).
    \OURMETHOD-H and \OURMETHOD-6 are trained on human images and all of the data.}
    \label{tab:qualitative_results_by_type}
\end{table}

%% file: src/figure/bad_crop_examples.tex
\begin{figure}[t]
    \centering
    \includegraphics[width=\columnwidth]{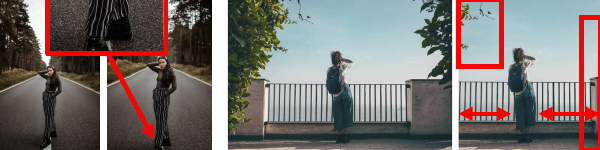}
    \caption{{\bf Examples of crops with subtle mistakes} (input on left; crop on right).
    First pair: the crop cuts through the subject's feet.
    Second pair: the crop leaves clutter on the edges and places the subject neither centered for left-right symmetry nor at a third, resulting in an unbalanced image.
    }
    \label{fig:bad_crop_examples}
\end{figure}

%% file: src/figure/condition.tex
\begin{figure}[t]
    \centering
    \includegraphics[width=\columnwidth]{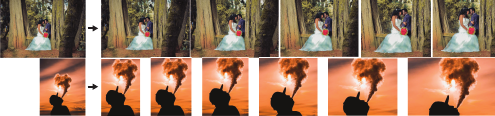}
    \caption{{\bf Conditional cropping model.}
    We are able to sample crop variations by varying the ``area'' (top) and ``aspect ratio'' (bottom) conditioning at inference time.}
    \label{fig:conditional_cropper}
\end{figure}

%% file: src/discussion.tex
\section{Discussion \& Conclusion}
\label{sec:discussion}

We have demonstrated that it is possible to equal or surpass fully-supervised performance on subject-aware image cropping using a weakly-supervised approach that requires only stock photos and a pre-trained generative model.

\OURMETHOD has its limitations.
Training and evaluating on generic data (unknown or arbitrary subjects) is difficult because of the distribution mismatch between professional images in Unsplash and the cropping datasets created by crowd-sourcing~\cite{fcdb,flms}.
Methods to calibrate Unsplash to these distributions could improve performance but are not a focus of this paper.
\OURMETHOD's pseudo-labels are sparse (1-per-image) and do not reproduce the dense crop-ranking annotations in GAICD, CPC, and SACD.
Other composition issues such as parallax and occlusions cannot be fixed by cropping alone.
Generative methods show promise for more advanced tasks such as searching for good perspectives in a NeRF~\cite{nerfwild} or a 3D-transformed photo~\cite{3dkenburns}.

Pairing inputs to labels is a common approach to learning, and obtaining both often requires an expensive annotation process.
Weakly-supervised learning typically relies on lower-quality-but-cheap labels to generalize and, by itself, is often not competitive with supervised approaches --- e.g., on ImageNet classification~\cite{imagenet}.
We have shown an instance where generative foundation models~\cite{stablediffusion} can invert this norm and convert plentiful, expert \emph{labels} (i.e., finished products) into otherwise difficult-to-obtain \emph{inputs}.
This has implications for other learning problems where obtaining complete training data is hard.
As the capabilities of image and text foundation models advance, we anticipate this paradigm will become a viable data creation strategy for other applications as well.

%% file: src/ack.tex
\section*{Acknowledgements}
This work is supported by gifts from Adobe and Meta as well as computing support from the Stanford Institute for Human-Centered Artificial Intelligence (HAI).
We thank Unsplash for providing access to their stock image dataset and the photographers who contributed their work to the Unsplash platform.
We also thank the anonymous reviewers for their helpful comments and feedback.

%% file: src/ethics.tex
\section*{Ethics Statement}

All of the images needed to reproduce this paper are publicly accessible on the Unsplash platform, under the terms of the Unsplash license (https://unsplash.com/license).
The images are designated by Unsplash for AI and academic use (https://unsplash.com/data).
Please refer to Unsplash for the full terms and conditions and access to image data.

There is growing ethical discussion around AI trained on public image data.
We use Unsplash images because they are highly moderated by Unsplash for artistic merit; inappropriate and offensive material; and data provenance.
Our methods are not limited to Unsplash images --- other professional stock image platforms, such as Adobe Stock, could also provide high-quality images that are suitable for training \OURMETHOD.
We encourage users to responsibly source their stock images according to the license terms of the image provider.

%% file: src_supp/supp_result.tex
\section{Additional Results and Ablations}
\label{sec:supp_additional_results}

\input{src_supp/supp_tab_violation}
\input{src_supp/supp_tab_ablation}
\input{src_supp/supp_tab_extra}

We include additional results omitted from the main paper.
Fig.~\ref{fig:supp_example_unsplash} and Fig.~\ref{fig:supp_example_fcdb} show cropping results on images from Unsplash~\cite{unsplash} and FCDB~\cite{fcdb}, respectively.
Fig.~\ref{fig:supp_conditional} shows additional examples by \OURMETHOD-C, our cropping model trained with area and aspect ratio conditioning.
This rest of the section is organized as follows.
\begin{itemize}
    \item \S\ref{sub:throughput_cost} analyzes the cost of our image generation pipeline.

    \item \S\ref{sub:full_qualitative} provides our full qualitative results on our five evaluation criteria introduced in \S\ref{sub:qualitative_eval} of the main text.

    \item \S\ref{sub:subject_awareness} -- \S\ref{sub:training_data_size} ablate subject awareness, data filtering, and training dataset size in order to better understand the performance of \OURMETHOD.

    \item \S\ref{sub:additional_modalities} and \S\ref{sub:model_arch_variations} modify our cropping model with additional inputs and different CNN architectures to investigate the extent that these reasonable deviations to model architecture affect performance.

    \item \S\ref{sub:generic_images} compares \OURMETHOD on generic images in FCDB~\cite{fcdb}, without the human-centric focus and subject-awareness.

    \item \S\ref{sub:crop_ranking} compares \OURMETHOD on human-centric crop ranking in GAICD~\cite{gaic}.
\end{itemize}
\noindent We also provide examples of generated images from our pipeline.
    Figs.~\ref{fig:supp_good_outpaint},~\ref{fig:supp_no_blip}, and~\ref{fig:supp_filter} show outpainted images produced by our pipeline and used for training; images outpainted without text-conditioning; and images that are discarded by filtering, respectively.
We hope that these additional results will provide useful baselines and commentary for future research on weakly-supervised image cropping.

\subsection{Computational Cost of Image Generation}
\label{sub:throughput_cost}

The most expensive stage of our pipeline is the generation of outpainted images.
This is dominated by Stable Diffusion~\cite{stablediffusion} (approximately 2 seconds per image on a NVIDIA V100 GPU~\cite{v100} at 512$\times$512 with 50 de-noising steps).
The other pre-trained models that we use for instance segmentation, YOLOv8x~\cite{yolov8}, and image captioning, BLIP-2 (6.7B)~\cite{blip2}, take 26 and 380 ms per image, respectively.
The overhead of our data quality filtering using the CNN and subject heuristic is negligible (2K images per second).

Our pipeline is trivially parallelizable across GPUs and machines, and costs approximately \$60 per 100K images, using spot VMs, making it both extremely fast and low-cost, even compared to crowd-sourcing annotations.


\subsection{Full Qualitative Results}
\label{sub:full_qualitative}

In the main results, we reported the number of cropping mistakes by method and category, aggregated by the five criteria: {\it Does the crop:
\begin{enumerate}
    \item cut unnaturally through the subject?
    \item cut unnaturally through the scene (e.g., other objects)?
    \item have too much or too little negative space?
    \item have unnecessary clutter around the edges?
    \item lack balance (e.g., missing symmetries, rule-of-thirds)?
\end{enumerate}}
\noindent Fig.~\ref{fig:supp_criteria} provides full definitions and examples of these criteria and their application to cropped images.

While these criteria still require an expert photographer to judge, they reduce the subjectivity of qualitative evaluation to key technical aspects of the image --- which can be consistently be found in reference books (such as~\cite{popphotobook,stunningbook}).
This assessment was performed by one of the paper authors, a photography domain expert, on 600 unique images (100 per subject category; 3,600 crops from the six methods) in a  blinded experiment, with all method names withheld and their ordering randomized.\footnote{Prior work~\cite{cacnet} also assessed the quality of cropped images using coarse `good', `normal', and `bad' buckets, but because our subject-aware task definition is more restrictive, we can apply a more focused set of technical criteria from the photography literature.}

We report the number of violations by each criterion in Tab.~\ref{tab:supp_qualitative}.
VFN~\cite{vfn}, the prior weakly-supervised method, performs very poorly in comparison to \OURMETHOD and the supervised HCIC~\cite{hcic} and CACNet~\cite{cacnet} baselines; performance is especially bad on cropping through the subject (V1) and negative space (V3), often by a factor of 2-4$\times$ or more.
We are unable to compare to SAC-Net~\cite{sacd} due to code not being available at time of submission.
On all subjects except horses, \OURMETHOD produced fewer unnatural cuts through the subject (V1) than HCIC, the second best.
\OURMETHOD also is similar or better than HCIC at managing negative space (V3) in the human, cat, dog, bird, and car images.
However, \OURMETHOD falls behind HCIC on V2 and V4, relating to contextual objects and clutter on the edges.
We believe that HCIC is able to better avoid these mistakes having learned on CPC~\cite{cpc} using their content-preservation scheme.
By contrast, \OURMETHOD regresses crops directly, and a crop with and without edge clutter can be very similar in L1 distance due to the difference being only a few pixels.
While our subject boundary loss penalizes unnatural cuts through the subject, defining such a loss over generic backgrounds and context is more challenging.
Horses are a challenging subject for \OURMETHOD, the number of initial stock images is smaller by a factor of 3x (2.2K) than the next smallest category, cats (6.7K).
In this case, \OURMETHOD-H trained on human images (22$\times$ more data), performs better than \OURMETHOD\footnote{Horse images may still contain humans such as riders.}.
On the other hand, we note that \OURMETHOD-H is worse than \OURMETHOD targeted for birds, suggesting that given a sufficient domain gap (human vs. bird appearance) and more specialized training data, training on more (out-of-domain) images alone does not guarantee improved performance.

\OURMETHOD-6 makes the fewest violations overall and suggests benefit from training on more diverse data.
While \OURMETHOD-H has already shown competitive results on subject-aware cropping problem posed by~\cite{sacd} (unconstrained by subject type), \OURMETHOD-6 suggests that one could automatically construct a generated dataset to better match a more generic distribution of categories by enumerating a small set of object classes.

\subsection{Ablation of Subject-Awareness}
\label{sub:subject_awareness}
Tab.~\ref{tab:supp_additional_experiments}a shows the importance of subject-awareness on the subject-aware benchmarks: the human-centric images in FLMS~\cite{flms} and FCDB~\cite{fcdb} used by~\cite{hcic}; the SACD~\cite{sacd} dataset; and our 1000 annotated human portraits (Portrait1K) from Unsplash~\cite{unsplash}.
We ablate the subject boundary loss, use of subject information (the mask and bounding box), and subject-focused dataset construction (by training on generic data outpainted from Unsplash images without filtering for humans).
These ablations are cumulative; removing subject information also removes the subject boundary loss since its computation depends on the subject mask; using generic images (reflecting the full content distribution of Unsplash) means that the subject and its type, if any, are not known.

We find that removing the subject boundary loss and subject information leads to a drop in IoU and increase in Disp. The result is small (up to 0.02 IoU and 0.007 Disp), compared to the large up-to 0.16 IoU and 0.05 Disp advantage \OURMETHOD has over the prior weakly-supervised method, VFN~\cite{vfn}.
Training \OURMETHOD on generic images (not restricted to humans and with no estimated subject) leads to a slightly larger loss of performance (up to 0.04 IoU and 0.01 Disp).
This shows that our outpainting based method is still effective over VFN, but that  selection of relevant data and some subject-awareness (e.g., using subject masks produced as a by-product of data filtering) is beneficial when the task is to crop images with a defined subject.

\subsection{Ablation of Data Filtering}
\label{sub:data_filtering}
The quality of the outpainted dataset has a small impact on the final cropping performance ranging from 0.03 to 0.01 on IoU and 0.008 to 0.002 on Disp (Tab.~\ref{tab:supp_additional_experiments}b).
As with subject-awareness, \OURMETHOD outperforms VFN~\cite{vfn} by a large margin even without filtering.
The effect of filtering with the CNN classifier, $D_{quality}$, and the additional outpainted subject heuristic are similar; many of the images removed by the subject heuristic are also removed by the CNN classifier, and vice versa.
For example, in Fig.~\ref{fig:supp_filter}a (columns 1, 2 and 4), a tiled or composite image is very likely to have additional instances of the subject class.

In the context of this work, the two data filtering steps, as well as our use of the image captioning model~\cite{blip2}, are implementation details for operating the current Stable Diffusion~\cite{stablediffusion} model.
We expect that future text-to-image inpainting models will produce fewer composite images or images with redundant subjects and be more faithful to text and image conditioning.
Recent works such as~\cite{controlnet,crossattentioncontrol,collagediffusion} have explored additional control for such text-to-image diffusion models and
these approaches could eliminate the need for data filtering by preventing undesirable content from being generated.

\subsection{Ablation of Training Data Size}
\label{sub:training_data_size}

We study the impact of the number of stock images needed to train \OURMETHOD.
%
%
In Tab.~\ref{tab:supp_additional_experiments}c, we vary the number of images, fixing the category to humans.

There is a steady drop off in performance on the human-centric images of FLMS + FCDB and on SACD as the number of starting images is reduced to 10K and then 1K.
With only 100 images, the performance is similar to removing subject-awareness and training on the unfiltered images in Unsplash -- compare to generic images, Tab.~\ref{tab:supp_additional_experiments}a.
More image diversity is clearly beneficial, possibly due to the domain gap between training on Unsplash images and testing on these datasets.

The fall-off is less severe on Portrait1K, our 1000 labeled images from Unsplash. With 10K images, the IoU and Disp actually increase slightly, before falling off at 1000 images and fewer.
The initial lift from reducing from 52K to 10K is due to our default hyperparameters and training schedule being more suited to a dataset approximately $\frac{1}{5}$ of the human dataset's size (similar in size to that of the cat, dog, bird, and car data);
on Portrait1K, we observe a similar boost with the full 52K images when the training schedule is shortened by $\frac{4}{5}$.
For consistency, we keep the same set of hyper-parameters across subject classes when training \OURMETHOD.

Training \OURMETHOD on a dataset generated with only 100 images still outperforms VFN~\cite{vfn} (trained with 52K images).
This clearly demonstrates the value that dataset generation via outpainting provides.

\subsection{Effect of Additional Input Modalities}
\label{sub:additional_modalities}

Photographers take into account factors such as shape and perspective (of which distance is a property) when composing an image.
We consider whether additional input representations such estimated depth~\cite{zoedepth} and Canny detected edges~\cite{canny} that loosely approximate these factors can improve cropping performance.
To test this, we concatenate these modalities as an additional input channel to the CNN feature extractor.
Tab.~\ref{tab:supp_additional_experiments}d shows that directly concatenating these inputs does not provide consistent benefit on IoU and Disp metrics.
Note that we do not explore larger modifications to the model architecture beyond concatenation nor do we change the other learning hyper-parameters, since these architectural directions are orthogonal to the dataset generation focus of our paper.

Future works may try to incorporate these priors, and our experiment here is to inform that the naive solution does not provide obvious benefit.

\subsection{Does Model Architecture Matter?}
\label{sub:model_arch_variations}

We test two variations of \OURMETHOD's architecture to evaluate whether the CNN architecture used for feature extraction matters and to test a simpler model architecture also trained on \OURMETHOD's generated data.

Prior works~\cite{cacnet,gaic,hcic,globalview,transview,vfn} have used a variety of CNN architectures for feature extraction, including AlexNet~\cite{alexnet}, VGG-16~\cite{vgg}, and MobileNet-V2~\cite{mobilenetv2}.
The choice of CNN for feature extraction has a small impact on the final cropping performance (Tab.~\ref{tab:supp_additional_experiments}c),
with VGG-16 being slightly worse on the three datasets and MobileNet-V2 being slightly better on Portrait1K, but worse on the others.
Reducing the model from ResNet-50~\cite{resnet} to ResNet-18 lowers performance.

We implemented a simpler baseline \OURMETHOD (U-Net), which uses a U-net predict a binary mask, representing each pixel's inclusion or exclusion in the crop.
At inference time, a threshold is applied and the bounding box of the largest connected component is predicted as the crop.
(See \S\ref{sub:supp_model_variants} for details.)
The \OURMETHOD U-net model performs slightly better than the regular \OURMETHOD on human-centric images in FLMS + FCDB but worse on SACD and Portrait1K, our labeled set of 1000 human images in Unsplash (\S\ref{sub:supp_our_labels}).

Overall, in both experiments, the magnitude of variation is small compared to the lift over VFN~\cite{vfn}.
This suggests that the specific cropping model architecture used for \OURMETHOD is not as important as the generated data.

Note that we do not claim the CACNet~\cite{cacnet} inspired architecture that we use in the main paper a key contribution of the paper.
There is a large space of possible models (such as the U-net) that could have been used in the experimentation, with likely similar effects.
However, we do observe that the specific design choices of our main approach are easily amenable to extension with the subject boundary loss (\S\ref{sub:cropping_architecture} in the main text) and conditional control (\S\ref{sub:conditional_cropper} in the main text).

\subsection{Comparison on Generic Images}
\label{sub:generic_images}

We focused on the subject-aware cropping task in the main text.
\OURMETHOD can also be used for `generic' images, by omitting the subject mask and subject-type specific filtering during dataset generation.
In this situation, we train on a pseudo-label distribution that directly reflects Unsplash~\cite{unsplash}.

We show results in Tab.~\ref{tab:supp_generic_fcdb} for cropping on FCDB~\cite{fcdb}.
\OURMETHOD performs significantly better than VFN~\cite{vfn}; approaches the performance of supervised models trained using GAICD~\cite{gaic} and AVA~\cite{ava}; but lags behind models trained on CPC~\cite{cpc}.
The simpler U-Net variant of \OURMETHOD described in \S\ref{sub:model_arch_variations} performs slightly better on FCDB, possibly because it is less expressive and prone to over-fit to Unsplash.

The distribution mismatch between a stock image dataset and benchmark dataset, such as FCDB~\cite{fcdb}, is a key confound.
For example, artistic images such as textures, where the subject is a pattern, in Unsplash are unlikely to be helpful on a benchmark like FCDB.
Likewise, for subsets of FCDB such as landscape images (where there is no clearly segmentable subject), the composition is often subpar compared to stock images even with an `optimal' crop due to inattention to perspective when the image was captured.
For these genres, we believe that helping casual photographers find better perspectives at capture time is a useful and that learning what makes an artistic or dramatic perspective for landscape from stock images is an interesting direction for future work.

\subsection{Comparison on Crop Ranking Tasks}
\label{sub:crop_ranking}

Models that are trained on densely annotated crop ranking datasets, e.g., GAICD~\cite{gaic}, are often evaluated using ranking metrics.
In this formulation of the cropping problem, the training data includes multiple crops per image along with ranks and scores.
This is in contrast to FCDB~\cite{fcdb} and \OURMETHOD, where supervision is sparse and each image has only a single label.

We find that these sparse datasets are insufficient to reproduce the scores in GAICD.
Ranking well involves calibration of various intermediate-quality crops to crop scores.
Tab.~\ref{tab:supp_ranking} shows results by \OURMETHOD-R, a naive implementation of crop ranking using our generated datasets (see \S\ref{sub:supp_model_variants} for details).

\input{src_supp/supp_fig_extra_results}

%% file: src_supp/supp_tab_violation.tex
\begin{table}[tp]
    {\centering \small \setlength{\tabcolsep}{5.5pt}
    \begin{tabularx}{\columnwidth}{llcrrrrr}
        \toprule
        & \multicolumn{1}{c}{Method} & Weak sup
            & V1 & V2 & V3 & V4 & V5 \\
        \midrule
        (a) & \multicolumn{7}{l}{{\em Portrait (Human)}} \\
        \midrule
            & VFN & \checkmark & 31 & 13 & 6 & 6 & 2 \\
            & CACNet && 7 & 2 & 1 & 1 & 0 \\
            & HCIC (CPC) && 5 & 3 & 0 & 1 & 1 \\
            & \OURMETHOD & \checkmark & 2 & 4 & 0 & 5 & 2 \\
            & \OURMETHOD-H & \checkmark &
                \multicolumn{5}{c}{(same as \OURMETHOD)} \\
            & \OURMETHOD-6 & \checkmark & 1 & 3 & 0 & 4 & 0 \\
        \midrule
        (b) & \multicolumn{7}{l}{{\em Cat}} \\
        \midrule
            & VFN & \checkmark & 26 & 6 & 9 & 1 & 2 \\
            & CACNet && 6 & 1 & 3 & 0 & 1 \\
            & HCIC (CPC) && 3 & 1 & 4 & 1 & 2 \\
            & \OURMETHOD & \checkmark & 2 & 2 & 0 & 1 & 2 \\
            & \OURMETHOD-H & \checkmark & 2 & 0 & 0 & 3 & 3 \\
            & \OURMETHOD-6 & \checkmark & 0 & 0 & 0 & 2 & 1 \\
        \midrule
        (c) & \multicolumn{7}{l}{{\em Dog}} \\
        \midrule
            & VFN & \checkmark & 28 & 19 & 15 & 5 & 8 \\
            & CACNet && 4 & 1 & 7 & 2 & 1 \\
            & HCIC (CPC) && 7 & 1 & 7 & 1 & 0 \\
            & \OURMETHOD & \checkmark & 1 & 3 & 1 & 1 & 0 \\
            & \OURMETHOD-H & \checkmark & 1 & 3 & 0 & 2 & 0 \\
            & \OURMETHOD-6 & \checkmark & 0 & 0 & 1 & 2 & 1 \\
        \midrule
        (d) & \multicolumn{7}{l}{{\em Horse}} \\
        \midrule
            & VFN & \checkmark & 40 & 8 & 14 & 0 & 9 \\
            & CACNet && 13 & 4 & 5 & 1 & 6 \\
            & HCIC (CPC) && 3 & 0 & 2 & 1 & 2 \\
            & \OURMETHOD & \checkmark & 9 & 4 & 5 & 1 & 0 \\
            & \OURMETHOD-H & \checkmark & 2 & 4 & 3 & 0 & 1 \\
            & \OURMETHOD-6 & \checkmark & 2 & 2 & 0 & 0 & 1 \\
        \midrule
        (e) & \multicolumn{7}{l}{{\em Bird}} \\
        \midrule
            & VFN & \checkmark & 46 & 8 & 13 & 1 & 7 \\
            & CACNet && 10 & 1 & 5 & 2 & 1 \\
            & HCIC (CPC) && 4 & 0 & 4 & 0 & 5 \\
            & \OURMETHOD & \checkmark & 1 & 1 & 0 & 1 & 0 \\
            & \OURMETHOD-H & \checkmark & 3 & 2 & 4 & 1 & 4 \\
            & \OURMETHOD-6 & \checkmark & 0 & 0 & 0 & 0 & 1 \\
        \midrule
        (f) & \multicolumn{7}{l}{{\em Car}} \\
        \midrule
            & VFN & \checkmark & 26 & 7 & 9 & 3 & 4 \\
            & CACNet && 1 & 2 & 6 & 0 & 2 \\
            & HCIC (CPC) && 0 & 0 & 0 & 0 & 3 \\
            & \OURMETHOD & \checkmark & 0 & 3 & 0 & 0 & 0 \\
            & \OURMETHOD-H & \checkmark & 0 & 2 & 2 & 1 & 0 \\
            & \OURMETHOD-6 & \checkmark & 2 & 3 & 1 & 2 & 2 \\
        \bottomrule
    \end{tabularx}
    }
    \caption{{\bf Full qualitative results.}
    Number of images with quality violation ($\downarrow$ is better) in 100 images sampled per class (a -- f).
    Refer to Fig.~\ref{fig:supp_criteria} for detailed explanation of the evaluation criteria (V1 -- V5).
    \OURMETHOD is our method trained on generated data of the subject class; \OURMETHOD-H is trained on generated data for humans; and \OURMETHOD-6 is trained jointly on generated data from all six classes.
    Analysis of these results is provided in \S\ref{sub:full_qualitative}.
    These results are summarized in Tab.~\ref{tab:qualitative_results} and Tab.~\ref{tab:qualitative_results_by_type} of the main text.
    }
    \label{tab:supp_qualitative}
\end{table}

%% file: src_supp/supp_tab_ablation.tex
\begin{table*}[t]
    \centering
    {\small
    \begin{tabularx}{0.7\textwidth}{llcccccc}
        \toprule
        &&
            \multicolumn{2}{c}{FLMS+FCDB} & \multicolumn{2}{c}{SACD}& \multicolumn{2}{c}{Portrait1K} \\
        & \multicolumn{1}{c}{Method}
            & \multicolumn{1}{c}{IoU $\uparrow$} & \multicolumn{1}{c}{Disp $\downarrow$}
            & \multicolumn{1}{c}{IoU $\uparrow$} & \multicolumn{1}{c}{Disp $\downarrow$}
            & \multicolumn{1}{c}{IoU $\uparrow$} & \multicolumn{1}{c}{Disp $\downarrow$} \\
        \midrule
        & VFN (for reference)
            & 0.5783 & 0.1114
            & 0.6555 & 0.0775
            & 0.6222 & 0.0948 \\
        & \OURMETHOD
            & 0.7334 & 0.0687
            & 0.7301 & 0.0632
            & 0.7501 & 0.0612 \\
        \midrule
        (a) & \multicolumn{7}{l}{{\em Ablation of subject-awareness}} \\
            & w/o subject boundary loss
            & 0.7145 & 0.0724
            & 0.7207 & 0.0653
            & 0.7458 & 0.0615 \\
        & w/o subject information
            & 0.7202 & 0.0714
            & 0.7105 & 0.0695
            & 0.7429 & 0.0628 \\
        & w/ generic images
            & 0.6861 & 0.0812
            & 0.7037 & 0.0716
            & 0.7396 & 0.0643 \\
        \midrule
        (b) & \multicolumn{7}{l}{{\em Ablation of data filtering}} \\
            & w/o CNN-based filter
            & 0.7112 & 0.0734
            & 0.7181 & 0.0666
            & 0.7409 & 0.0632 \\
        & w/o CNN \& heuristic filter
            & 0.6987 & 0.0765
            & 0.7217 & 0.0656
            & 0.7398 & 0.0640 \\
        \midrule
        (c) & \multicolumn{7}{l}{{\em Ablation of training dataset size (number of stock images used for outpainting)}} \\
            & w/ 10000 images (19.2 \%)
                & 0.7174 & 0.0719
                & 0.7238 & 0.0658
                & 0.7640 & 0.0570 \\
            & w/ 1000 images (1.9 \%)
                & 0.7007 & 0.0756
                & 0.7215 & 0.0658
                & 0.7441 & 0.0613 \\
            & w/ 100 images (0.2 \%)
                & 0.6865 & 0.0780
                & 0.6918 & 0.0709
                & 0.7143 & 0.0655 \\
        \midrule
        (d) & \multicolumn{7}{l}{{\em Additional inputs to the cropping model}} \\
            & w/ depth~\cite{zoedepth}
                & 0.7226 & 0.0708
                & 0.7245 & 0.0649
                & 0.7505 & 0.0610 \\
            & w/ edges~\cite{canny}
                & 0.7205 & 0.0715
                & 0.7225 & 0.0661
                & 0.7477 & 0.0619 \\
        \midrule
        (e) & \multicolumn{7}{l}{{\em Different CNN feature extractors}} \\
        & ResNet-18
            & 0.7181 & 0.0726
            & 0.7237 & 0.0648
            & 0.7408 & 0.0640 \\
        & VGG-16
            & 0.7241 & 0.0706
            & 0.7247 & 0.0635
            & 0.7457 & 0.0623 \\
        & MobileNet-V2
            & 0.7207 & 0.0710
            & 0.7284 & 0.0633
            & 0.7565 & 0.0592 \\
        \midrule
        (f) & \multicolumn{7}{l}{{\em Different model architectures}} \\
        & U-Net (see \S\ref{sub:supp_model_variants} for details)
            & 0.7365 & 0.0683
            & 0.7195 & 0.0680
            & 0.7420 & 0.0652 \\
        \bottomrule
    \end{tabularx}
    }
    \caption{{\bf Ablations and additional experiments.}
    FCDB+FLMS is the human-centric test-set used by~\cite{hcic}.
    SACD is the subject-aware dataset from~\cite{sacd}.
    Portrait1K refers to our annotated human images from Unsplash (\S\ref{sub:supp_our_labels}).
    We provide the results for \OURMETHOD and VFN reported in the main paper for reference.
    Analysis of these results is provided in the corresponding sections of~\S\ref{sec:supp_additional_results}.
    At a high level, we find that:
    (a) Knowledge about the subject benefits subject-aware cropping.
    (b) Removing poorly outpainted images improves cropping model accuracy.
    (c) Reducing the number of stock images available for outpainting lowers performance, as there is less diversity in the dataset.
    (d) Additional inputs (such as depth or edge detections) to the cropping model do not provide clear benefits.
    (e) Different CNN feature extractors provide similar performance as the ResNet-50 used in our main results.
    (f) \OURMETHOD is not bound to a particular model architecture. The value from \OURMETHOD comes from dataset generation, and even a simpler U-Net baseline (implementation details in \S\ref{sub:supp_model_variants}) can realize the benefits of our data generation approach.
    }
    \label{tab:supp_additional_experiments}
\end{table*}

%% file: src_supp/supp_tab_extra.tex
\begin{table}[tp]
    {\centering \small \setlength{\tabcolsep}{2pt}
    \begin{tabularx}{\columnwidth}{lccrr}
        \toprule
        \multicolumn{1}{c}{Method} & \multicolumn{1}{c}{Trained on} & Weak sup
            & \multicolumn{1}{c}{IoU $\uparrow$} & \multicolumn{1}{c}{Disp $\downarrow$} \\
        \midrule
        A2RL~\cite{a2rl} & AVA & & 0.663 & 0.082 \\
        A3RL~\cite{a3rl} & AVA & & 0.696 & 0.077 \\
        VPN~\cite{cpc} & CPC & & 0.711 & 0.073 \\
        VEN~\cite{cpc} & CPC & & 0.735 & 0.072 \\
        ASM~\cite{asm} & CPC & & 0.749 & 0.068 \\
        GAIC~\cite{gaic} & GAICD & & 0.672 & 0.084 \\
        CGS~\cite{cgs} & GAICD & & 0.685 & 0.079 \\
        TransView~\cite{transview} & GAICD && 0.685 & 0.080 \\
        \cite{rankconsistentcrop} & GAICD && 0.686 & 0.078 \\
        \midrule
        VFN~\cite{vfn} & Unsplash & \checkmark & 0.450 & 0.147 \\
        \OURMETHOD & Unsplash & \checkmark & 0.654 & 0.090 \\
        \OURMETHOD (U-Net) & Unsplash & \checkmark & 0.670 & 0.086 \\
        \bottomrule
    \end{tabularx}
    }
    \caption{{\bf Evaluation on generic images in FCDB}~\cite{fcdb}.
    Apart from VFN~\cite{vfn} which we train on Unsplash, the reported numbers are from the original papers and~\cite{rankconsistentcrop}.}
    \label{tab:supp_generic_fcdb}
\end{table}

\begin{table}[tp]
    {\centering \small \setlength{\tabcolsep}{2.8pt}
    \begin{tabularx}{\columnwidth}{lccrrr}
        \toprule
        \multicolumn{1}{c}{Method} & \multicolumn{1}{c}{Trained on} & Weak sup
            & \multicolumn{1}{c}{$\overline{SRCC}$ $\uparrow$}
            & \multicolumn{1}{c}{$\overline{Acc_5}$ $\uparrow$}
            & \multicolumn{1}{c}{$\overline{Acc_{10}}$ $\uparrow$} \\
        \midrule
        VFN & GAICD & & \fromhcic0.648 & \fromhcic41.3 & \fromhcic60.2 \\
        HCIC & GAICD & & \fromhcic0.795 & \fromhcic59.7 & \fromhcic77.0 \\
        \midrule
        VFN & Flickr & \checkmark & \fromhcic0.332 & \fromhcic10.1 & \fromhcic21.1 \\
        VFN & Unsplash & \checkmark & 0.203 & 13.3 & 19.1 \\
        \OURMETHOD-R & Unsplash & \checkmark & 0.446 & 23.1 & 38.5 \\
        \bottomrule
    \end{tabularx}
    }
    \caption{{\bf Comparison on the 50 human-centric test images in GAICD.}
    \fromhcic~indicates results reported by~\cite{hcic}; refer to their paper for the full table of baselines.
    As noted in \S\ref{sub:crop_ranking}, the sparse supervision generated by \OURMETHOD is poorly calibrated to the ranking labels in GAICD.
    Our version of \OURMETHOD (\OURMETHOD-R), modified for ranking performs better than VFN, which also does not have access to score and rank labels (e.g., being  trained on Flickr or Unsplash), but results are poor compared to models that directly train on GAICD.
    }
    \label{tab:supp_ranking}
\end{table}

\begin{table}[tp]
    {\centering \small \setlength{\tabcolsep}{6pt}
    \begin{tabularx}{\columnwidth}{lrr}
        \toprule
        \multicolumn{1}{c}{Method}
            & \multicolumn{1}{c}{\# Params}
            & \multicolumn{1}{c}{Inference time (ms)} \\
        \midrule
        HCIC~\cite{hcic} & 19.47M & *7.8 \\
        CACNet~\cite{cacnet} & 18.93M & 3.4 \\
        \OURMETHOD & 24.93M & 5.7 \\
        \bottomrule
    \end{tabularx}
    }
    \caption{{\bf Comparison between models: \OURMETHOD and the baselines.}
    Cropping model complexity and inference time are not a key priority of this paper, but we include these details for completeness.
    We measured inference time per image on a single NVIDIA RTX A5000 GPU~\cite{a5000},
    except for HCIC (*) which is their reported inference time (128 FPS or 7.8 milliseconds), computed on a slightly more powerful RTX 3090.
    We were unable to reproduce HCIC's performance using the available code.
    \OURMETHOD is slower than CACNet~\cite{cacnet} by 1.7$\times$, but faster than HCIC~\cite{hcic} by 0.37$\times$ (despite a less powerful GPU).
    Both \OURMETHOD and CACNet directly regress a crop, while HCIC ranks based on a set of candidates.
    }
    \label{tab:supp_inference_speed}
\end{table}

%% file: src_supp/supp_fig_extra_results.tex
\begin{figure*}[p]
    \centering
    \includegraphics[width=\textwidth]{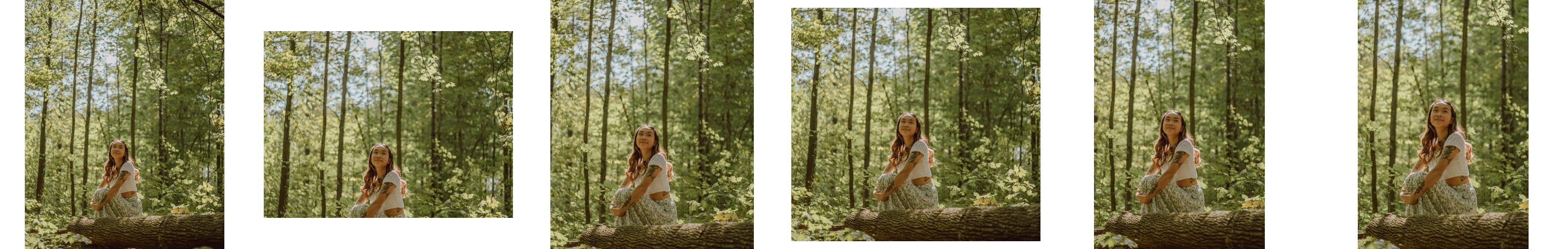}

    \vspace{0.05in}

    \includegraphics[width=\textwidth]{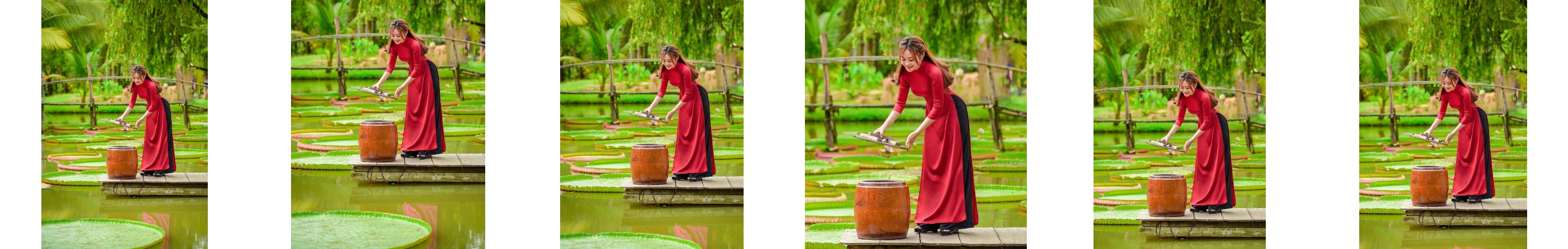}

    \vspace{0.05in}

    \includegraphics[width=\textwidth]{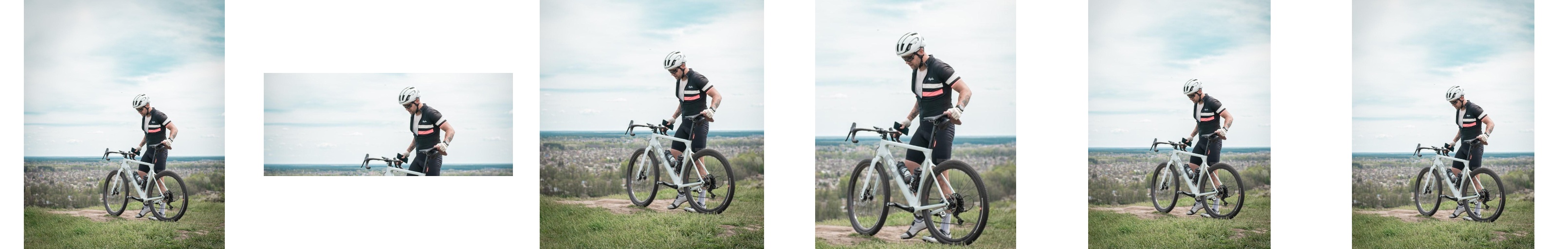}

    \vspace{0.05in}

    \includegraphics[width=\textwidth]{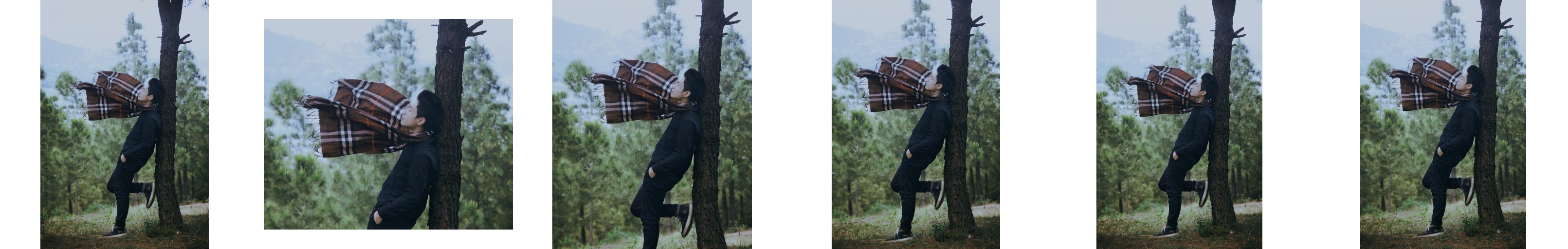}

    \vspace{0.05in}

    \includegraphics[width=\textwidth]{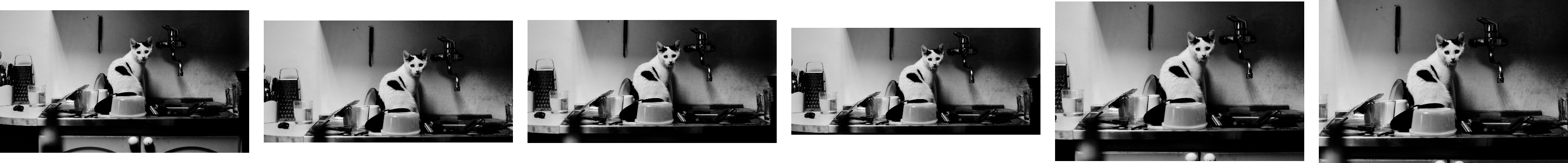}

    \vspace{0.05in}

    \includegraphics[width=\textwidth]{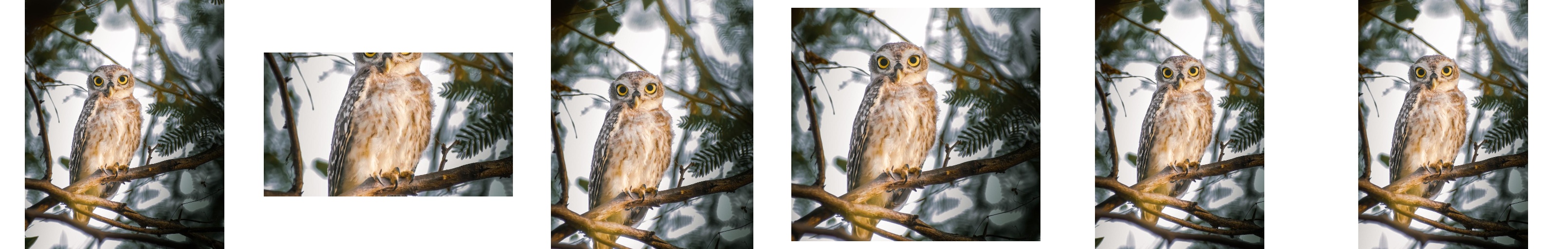}

    \vspace{0.05in}

    \includegraphics[width=\textwidth]{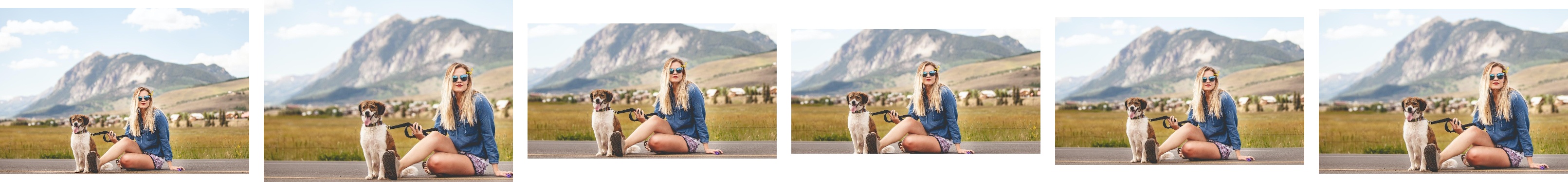}

    \vspace{0.05in}

    \includegraphics[width=\textwidth]{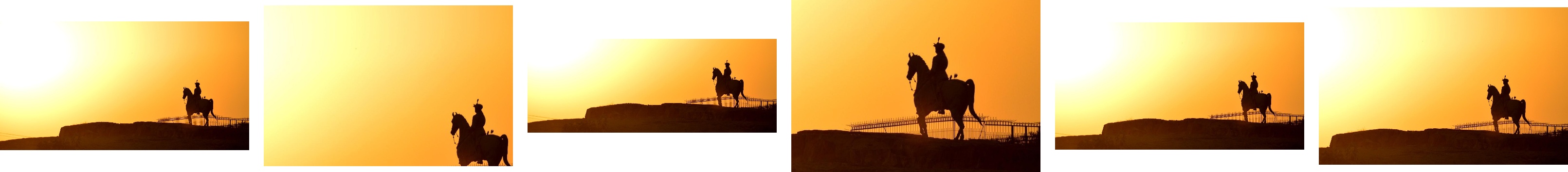}

    \vspace{0.1in}

    \makebox[0.162895\textwidth][c]{Input image}
    \makebox[0.162895\textwidth][c]{VFN}
    \makebox[0.162895\textwidth][c]{CACNet}
    \makebox[0.162895\textwidth][c]{HCIC}
    \makebox[0.162895\textwidth][c]{\OURMETHOD}
    \makebox[0.162895\textwidth][c]{\OURMETHOD-6}

    \caption{{\bf Example crops on Unsplash.}~\cite{unsplash}
    \OURMETHOD and \OURMETHOD-6 are trained on generated images of the subject category and images of all six studied categories (listed in \S\ref{sec:dataset} of the main text), respectively. \OURMETHOD produces similar quality results to (supervised) HCIC while VFN performs poorly.
    }
    \label{fig:supp_example_unsplash}
\end{figure*}

\begin{figure*}[p]
    \centering
    \includegraphics[width=\textwidth]{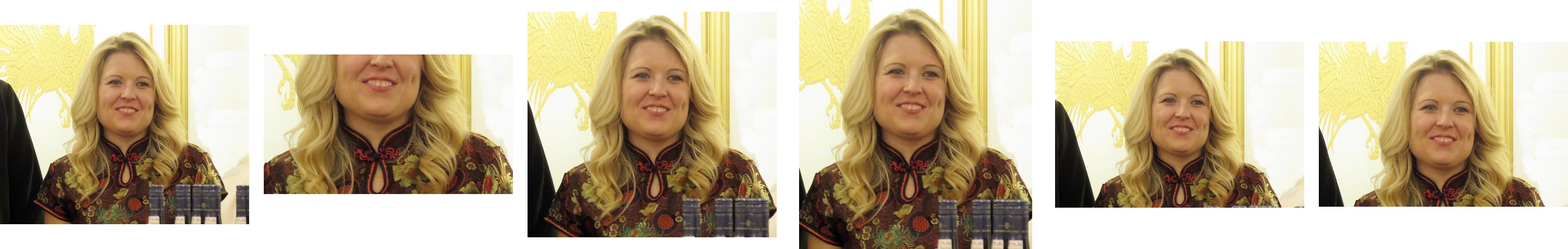}

    \vspace{0.05in}

    \includegraphics[width=\textwidth]{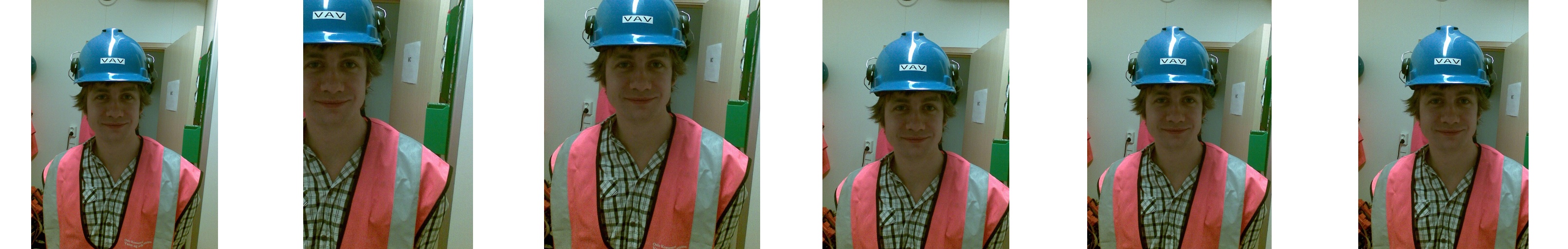}

    \vspace{0.05in}

    \includegraphics[width=\textwidth]{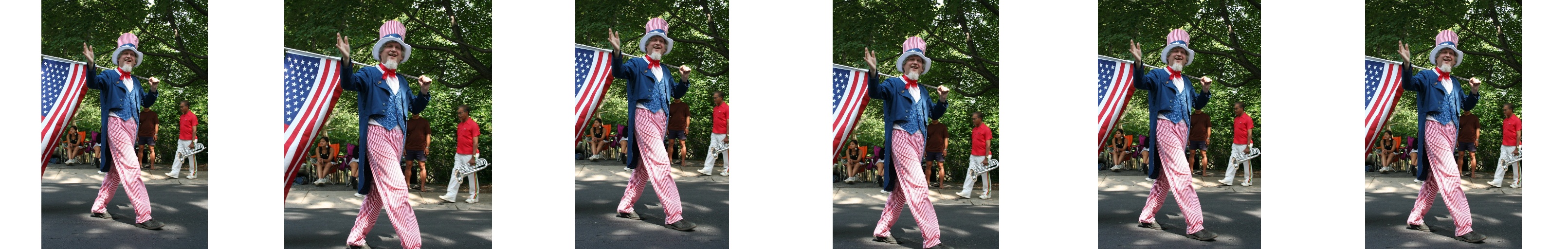}

    \vspace{0.05in}

    \includegraphics[width=\textwidth]{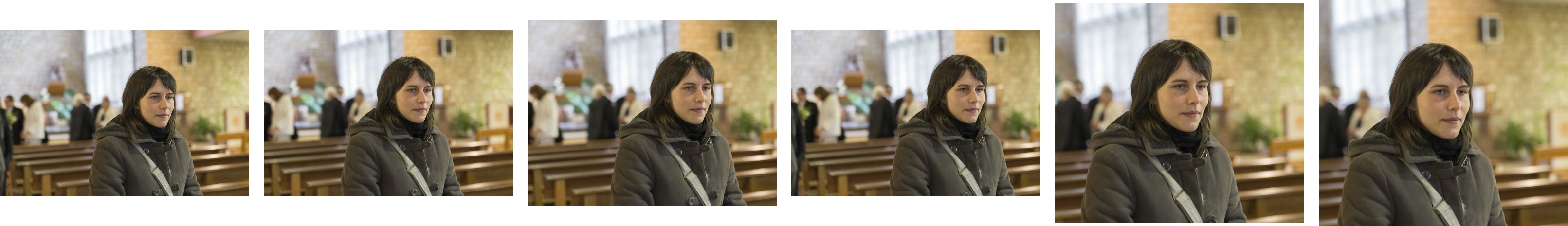}

    \vspace{0.05in}

    \includegraphics[width=\textwidth]{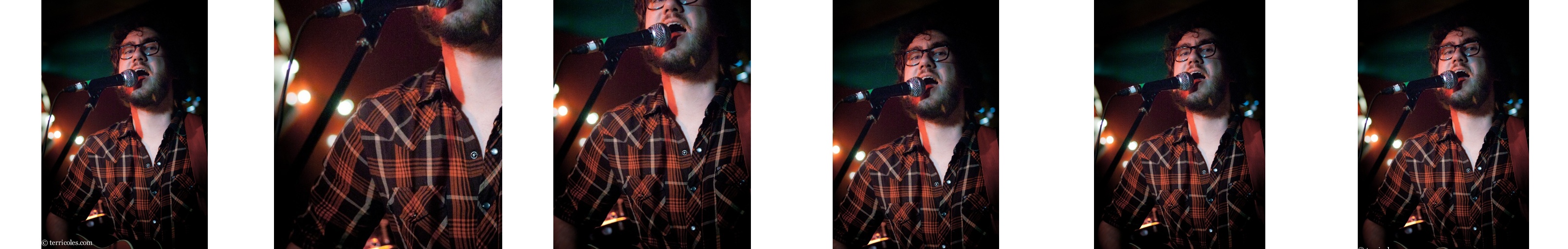}

    \vspace{0.05in}

    \includegraphics[width=\textwidth]{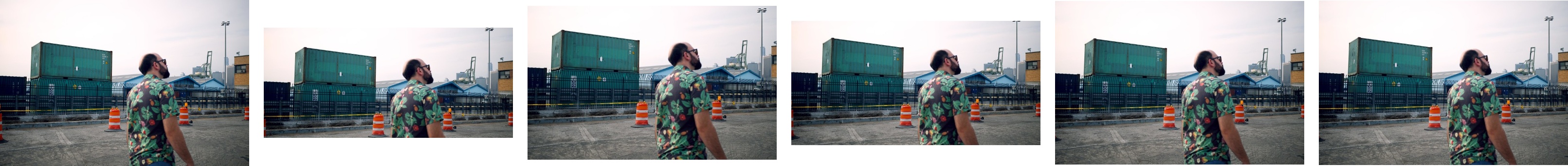}

    \vspace{0.05in}

    \includegraphics[width=\textwidth]{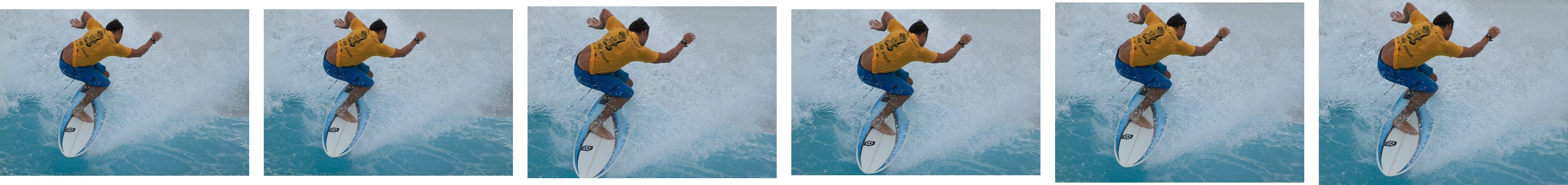}

    \vspace{0.1in}

    \makebox[0.162895\textwidth][c]{Input image}
    \makebox[0.162895\textwidth][c]{VFN}
    \makebox[0.162895\textwidth][c]{CACNet}
    \makebox[0.162895\textwidth][c]{HCIC}
    \makebox[0.162895\textwidth][c]{\OURMETHOD}
    \makebox[0.162895\textwidth][c]{\OURMETHOD-6}

    \caption{{\bf Example crops for human-centric images in FCDB.}~\cite{fcdb}
    \OURMETHOD and \OURMETHOD-6 are trained on generated images of humans and images of all six studied categories (listed in \S\ref{sec:dataset} of the main text), respectively.
    \OURMETHOD produces similar quality results to (supervised) HCIC while VFN performs poorly.
    }
    \label{fig:supp_example_fcdb}
\end{figure*}

\begin{figure*}[p]
    \centering
    \begin{subfigure}[t]{\textwidth}
        \centering
        \includegraphics[width=\textwidth]{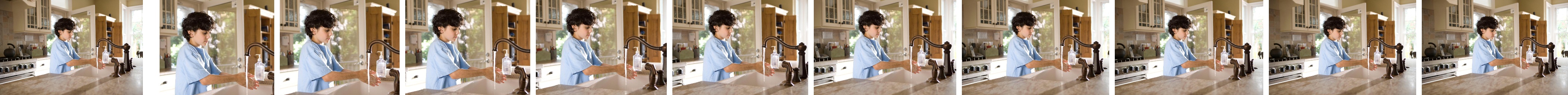}

        \vspace{0.05in}

        \includegraphics[width=\textwidth]{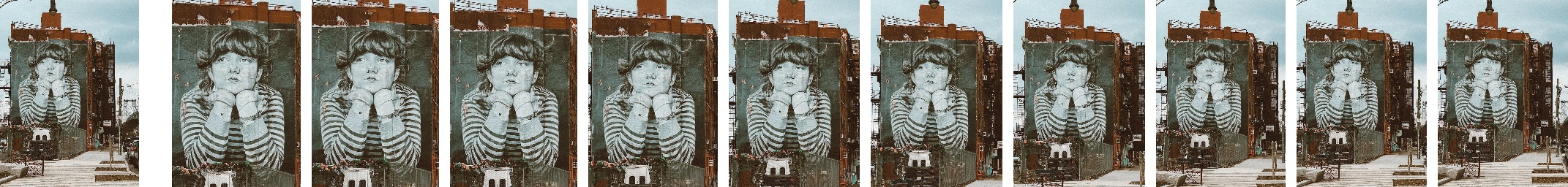}

        \vspace{0.05in}

        \includegraphics[width=\textwidth]{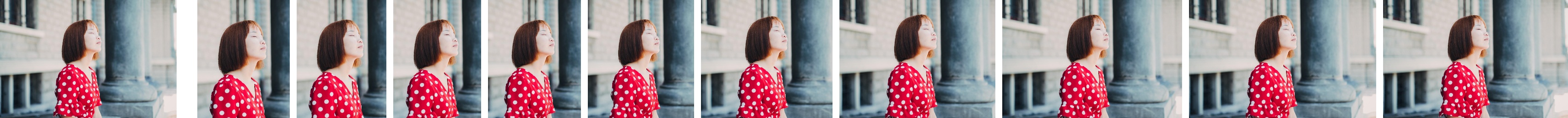}

        \vspace{0.05in}

        \includegraphics[width=\textwidth]{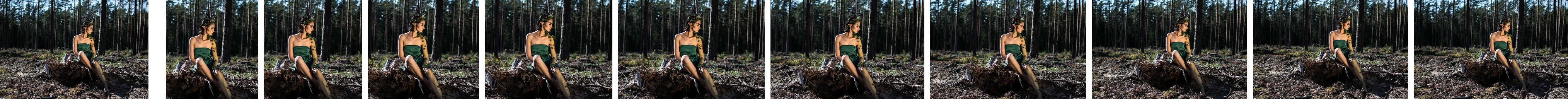}
        \caption{{\em Conditioning on area.} (Tight to loose)}
    \end{subfigure}

    \vspace{1em}

    \centering
    \begin{subfigure}[t]{\textwidth}
        \centering
        \includegraphics[width=\textwidth]{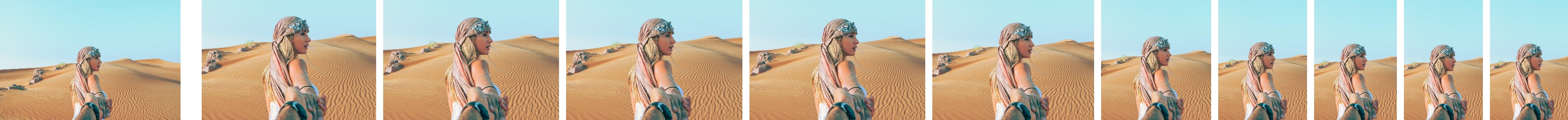}

        \vspace{0.05in}

        \includegraphics[width=\textwidth]{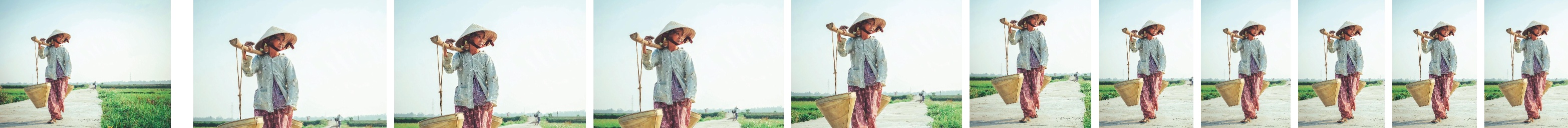}

        \vspace{0.05in}

        \includegraphics[width=\textwidth]{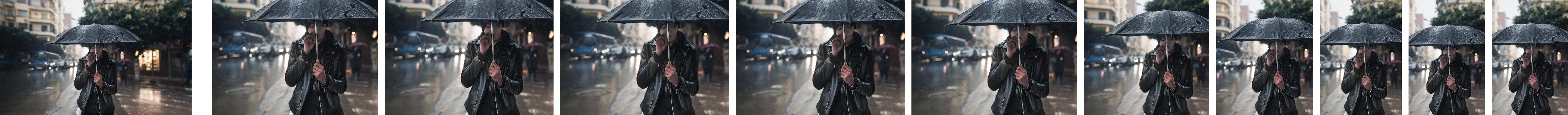}

        \vspace{0.05in}

        \includegraphics[width=\textwidth]{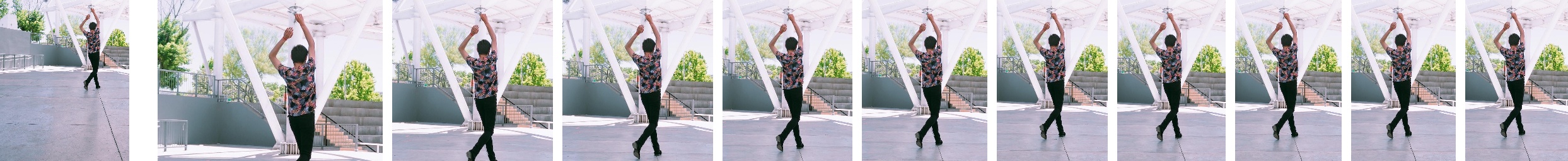}
        \caption{{\em Conditioning on aspect ratio.} (Wide to tall)}
    \end{subfigure}
    \caption{{\bf Additional examples of cropping with conditioning (\OURMETHOD-C).}
    Original image on the left.
    By sweeping the value of the conditioning signal, (a) from 0.1 to 1 for area and (b) from 16:9 to 9:16 for aspect ratio, we are able to sample crops of different tightness.
    Note that extreme values can cause unnatural crops such as in the last row of (b), where the portrait orientation of the starting image provides limited room for a wide, landscape crop.
    \OURMETHOD-C also does not enforce exact adherence to the conditioning at inference time; the conditioning is a continuous input signal akin to a hint to the model that is considered jointly with the image content.
    }
    \label{fig:supp_conditional}
\end{figure*}

\begin{figure*}[p]
    \centering
    \begin{subfigure}[t]{\textwidth}
        \centering
        \includegraphics[width=0.73\textwidth]{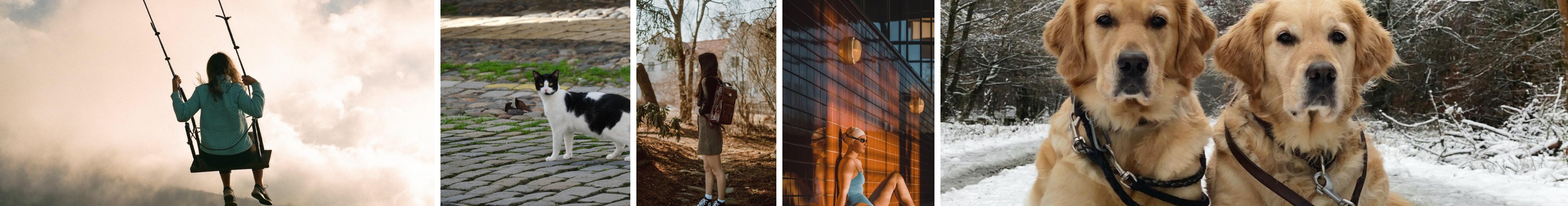}
        \caption{{\bf V1:} {\em Does the crop cut unnaturally through the subject?}
        A generally accepted rule of portrait photography is to never cut a person through a joint or through the chin.
        A common mistake by cropping models is to cut the subject slightly through the feet, ankles, or hands as these are often furthest from the subject center.
        We label these and more egregious errors as violations. For non-humans, we apply similar criteria for whether a cut through a subject is unnatural.
        In images with multiple instances of the subject category (e.g., multiple people) we assess cuts through any of the possible subjects.
        }
    \end{subfigure}

    \vspace{0.2em}

    \begin{subfigure}[t]{\textwidth}
        \centering
        \includegraphics[width=0.73\textwidth]{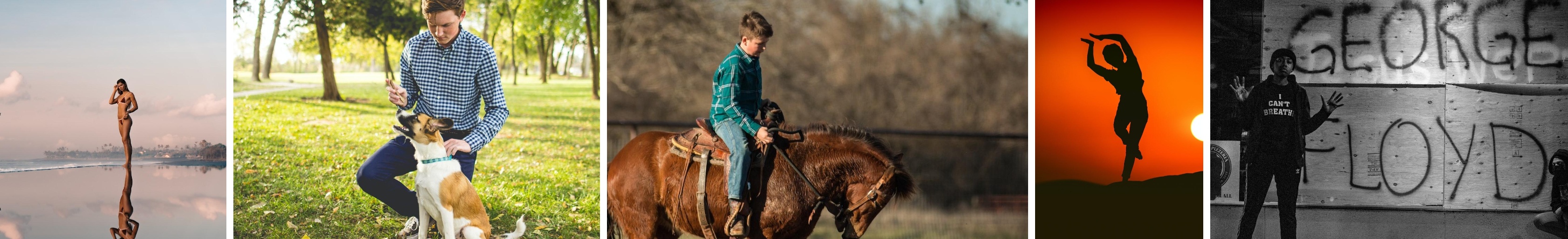}
        \caption{{\bf V2:} {\em  Does the crop cut unnaturally through the scene?}
        The scene can include other objects of interest in addition to the subject.
        Sometimes these are objects that the subject is interacting with.
        We consider it an error if a crop removes part of an object in an unnatural or distracting way.
        The example images crop the subject's reflection at the neck, the dog's feet, the horse's head, the sun, and a 2nd person in bottom right.
        }
    \end{subfigure}

    \vspace{0.2em}

    \begin{subfigure}[t]{\textwidth}
        \centering
        \includegraphics[width=0.73\textwidth]{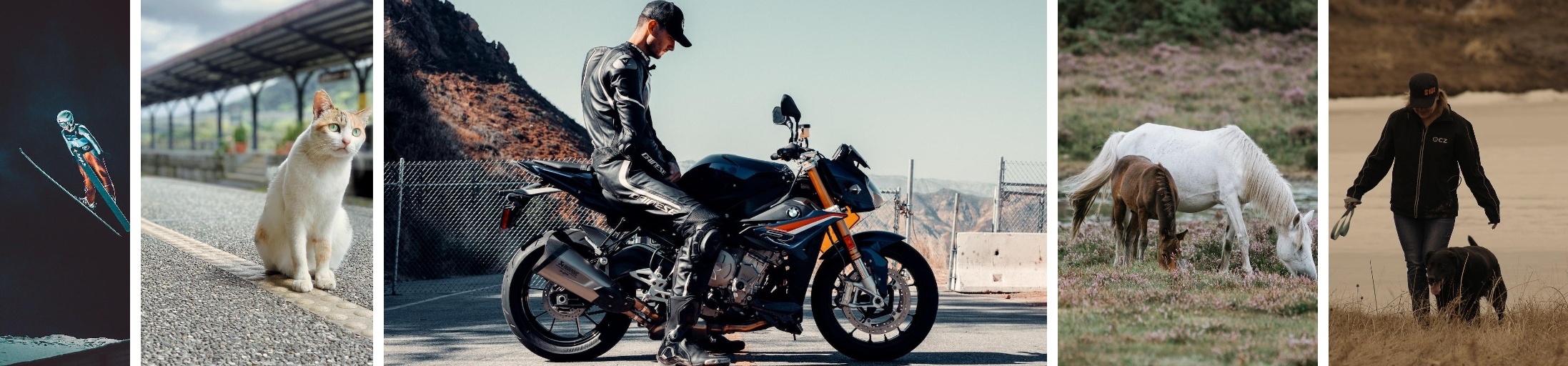}
        \caption{{\bf V3:} {\em Does the crop have too much or too little negative space?}
        Negative space is empty space around and between subjects in an image, with positive space being the space occupied by subjects.
        The presence of negative space is often used to draw attention to the subject.
        We label a crop as having too much or too little negative space if negative space is missing entirely (a very tight crop) or if there are unbalanced amounts of negative space between the sides of the image (e.g., a large amount of negative space horizontally, but a very tight crop vertically).
        Often more negative space is desired in the direction of the subject's gaze or movement~\cite{stunningbook}.
        }
    \end{subfigure}

    \vspace{0.2em}

    \begin{subfigure}[t]{\textwidth}
        \centering
        \includegraphics[width=0.73\textwidth]{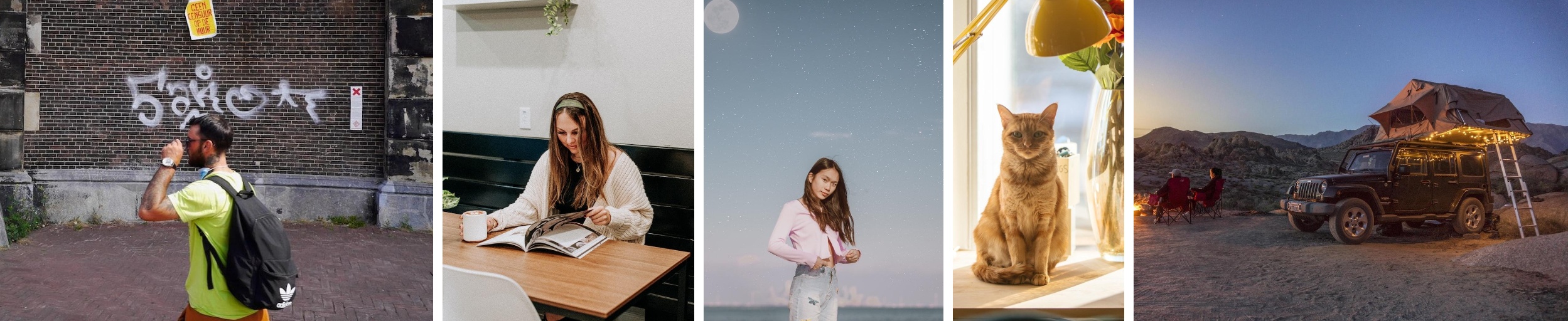}
        \caption{{\bf V4:} {\em Does the crop have unnecessary clutter around the edges?}
        Cropping can improve framing by removing distractors from an image.
        However, cropping the scene can also introduce distractors if salient objects or areas (e.g., bright areas, people) are only removed partially or placed on the edges of the image.
        We label an crop as having introduced clutter if there is a similar / better crop that would have removed or included the salient region.
        }
    \end{subfigure}

    \vspace{0.2em}

    \begin{subfigure}[t]{\textwidth}
        \centering
        \includegraphics[width=0.73\textwidth]{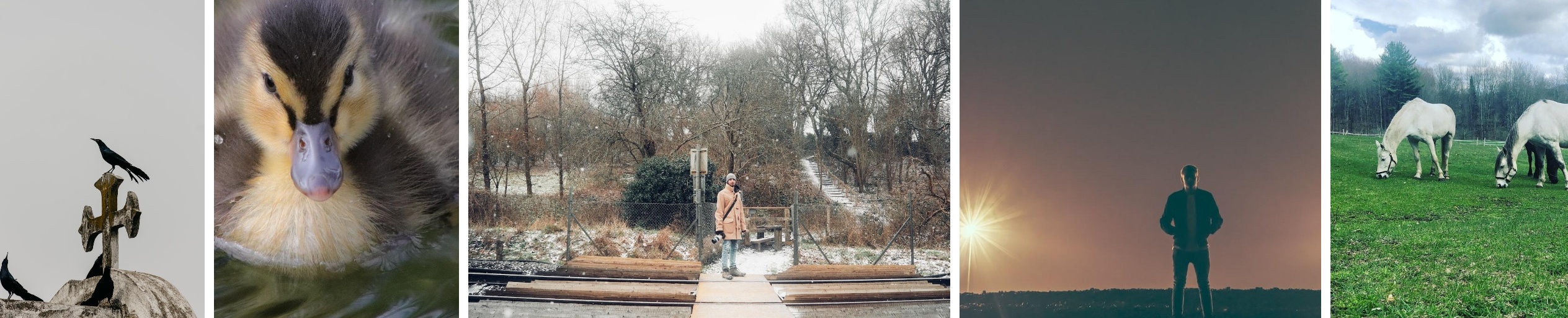}
        \caption{{\bf V5:} {\em Does the crop lack balance?}
        Positioning of areas of interest in the frame is an important part of  composition.
        The Rule-of-Thirds is often promoted as an alternative to placing a subject directly centered.
        Other common techniques include looking for symmetries or lines that lead into the image or arranging regions of interest to balance the sides of the image.
        For example, a framing might be nearly symmetric but slightly off from the axis of symmetry, or the framing may place salient content too far toward the edges.
        In the example images, there are more balanced compositions that can be achieved by slightly adjusting the crop.
        }
    \end{subfigure}

    \caption{{\bf Criteria for qualitative evaluation and example failures.}
    (The example crops are taken from the baselines and \OURMETHOD.)
    We use these guidelines to control the subjectivity of qualitative evaluation and to provide insight on the types of cropping errors made by the various models.
    Note that the criteria are not mutually exclusive.
    }
    \label{fig:supp_criteria}
\end{figure*}

\begin{figure*}[p]
    \centering
    \includegraphics[width=0.195\textwidth]{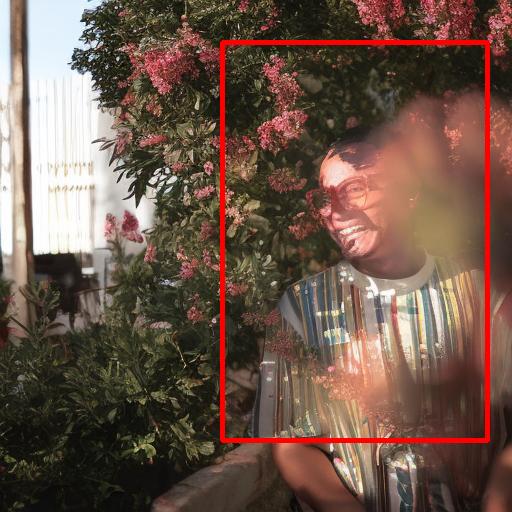}
    \includegraphics[width=0.195\textwidth]{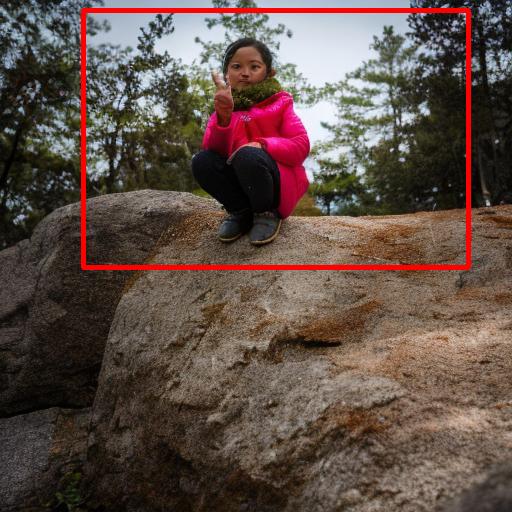}
    \includegraphics[width=0.195\textwidth]{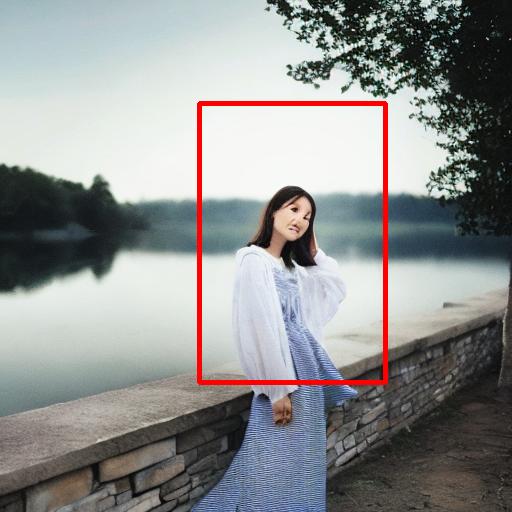}
    \includegraphics[width=0.195\textwidth]{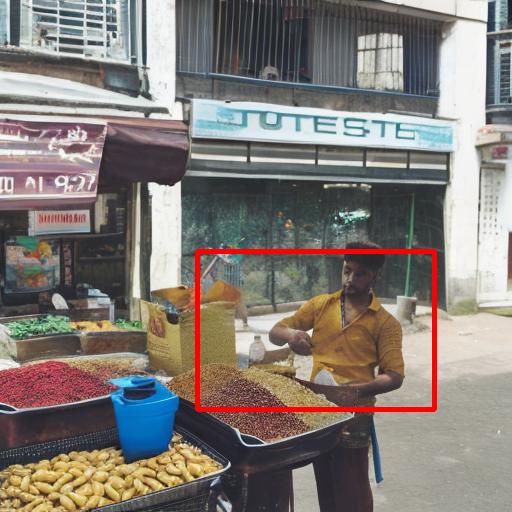}
    \includegraphics[width=0.195\textwidth]{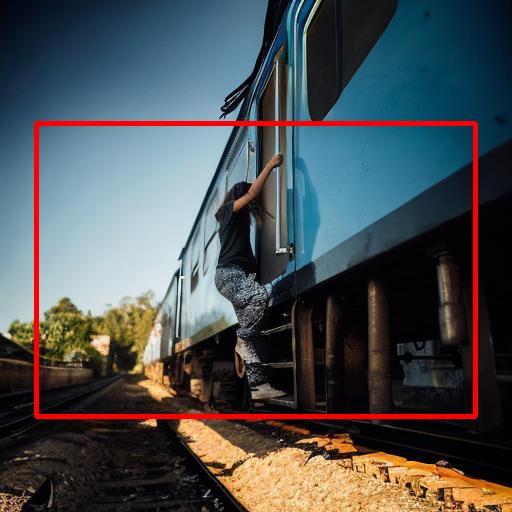}

    \vspace{0.1em}

    \centering
    \includegraphics[width=0.195\textwidth]{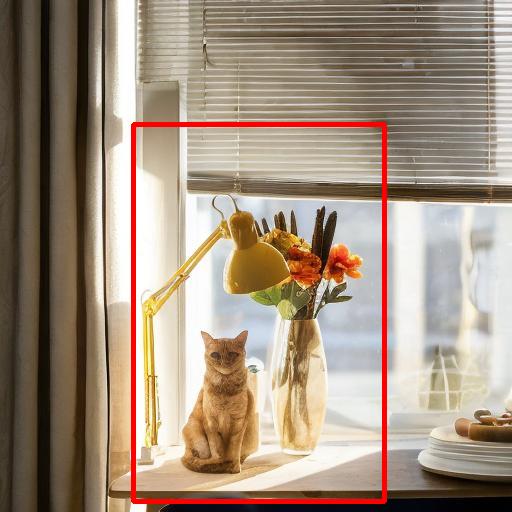}
    \includegraphics[width=0.195\textwidth]{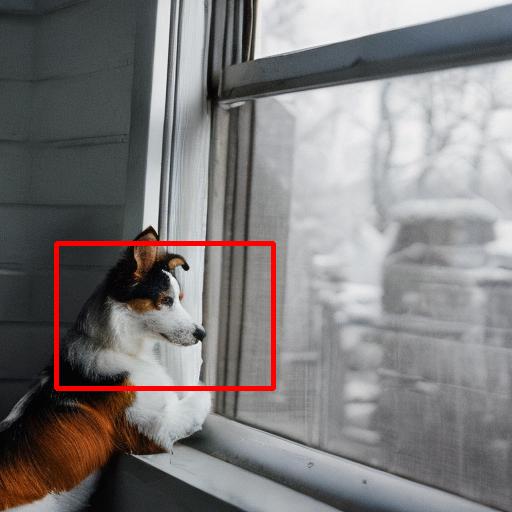}
    \includegraphics[width=0.195\textwidth]{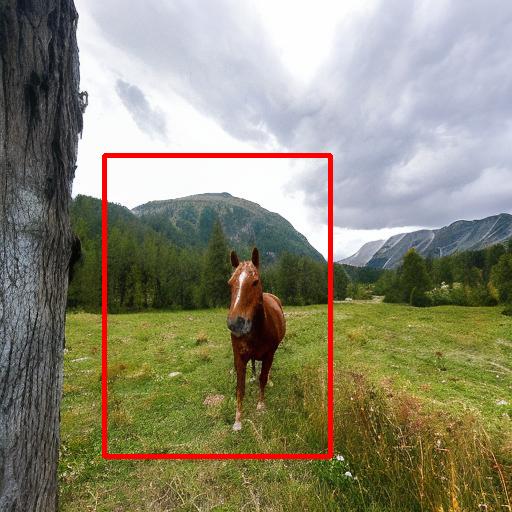}
    \includegraphics[width=0.195\textwidth]{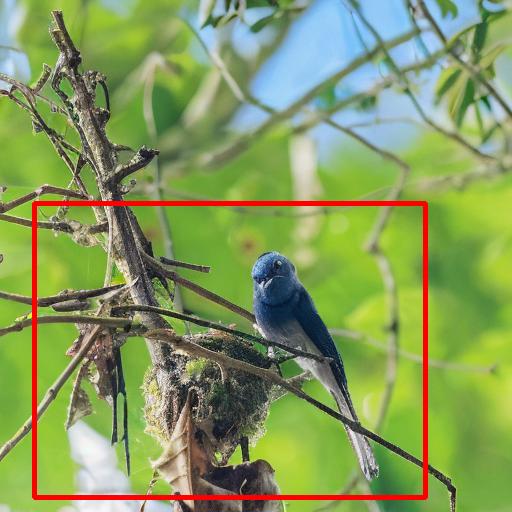}
    \includegraphics[width=0.195\textwidth]{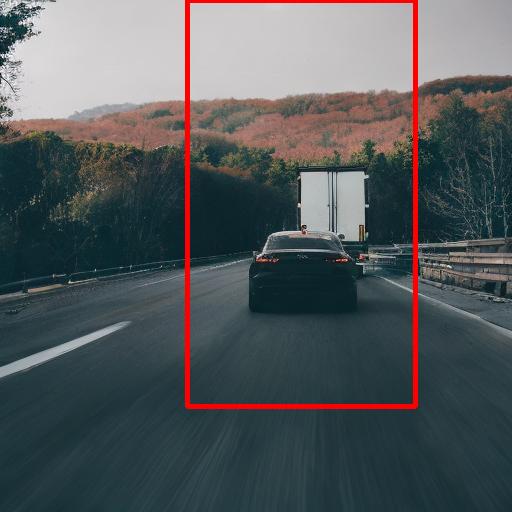}

    \caption{{\bf Outpainted images used to train \OURMETHOD.} A random sample.
    Red shows the original image.
    Stable Diffusion~\cite{stablediffusion} is able to produce a plausible, uncropped scene that a photographer might have seen.
    Contrast this with Fig.~\ref{fig:supp_filter}, which shows images that were discarded by our filtering and not used for training.
    }
    \label{fig:supp_good_outpaint}
\end{figure*}

\begin{figure*}[p]
    \centering
    \begin{subfigure}[t]{0.495\textwidth}
        \centering
        \includegraphics[width=0.49\textwidth]{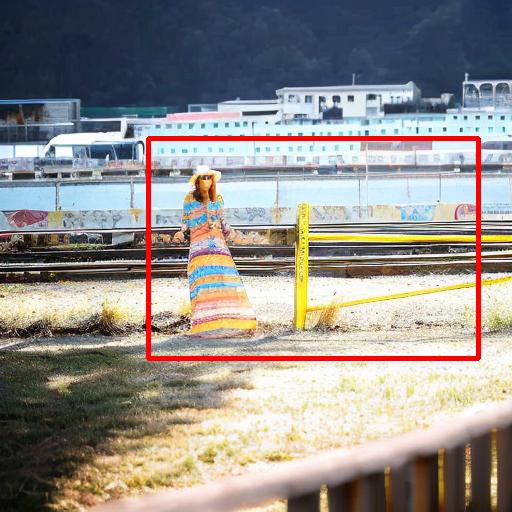}
        \includegraphics[width=0.49\textwidth]{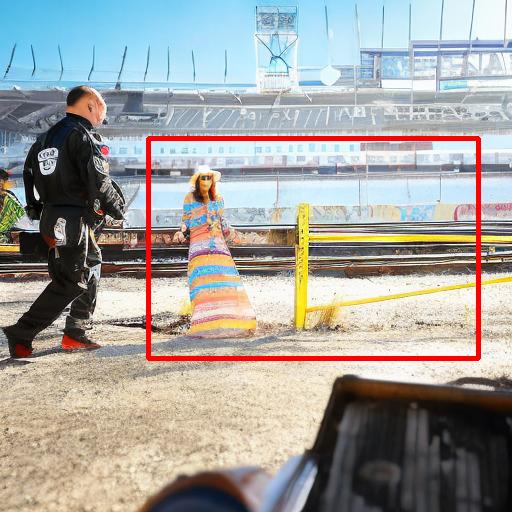}
        \caption{{\em ``a woman in a colorful dress sanding on a fence''}}
    \end{subfigure}
    \hfill
    \begin{subfigure}[t]{0.495\textwidth}
        \centering
        \includegraphics[width=0.49\textwidth]{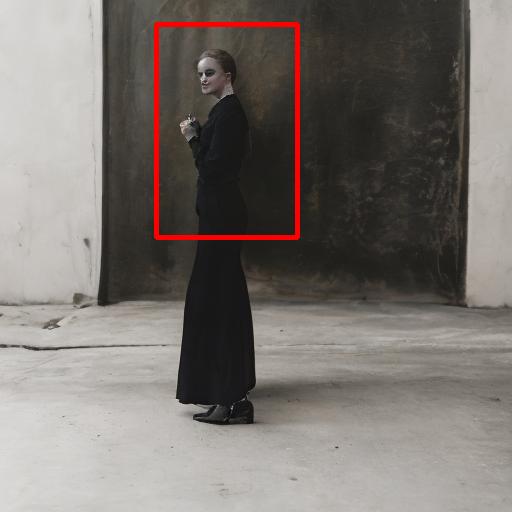}
        \includegraphics[width=0.49\textwidth]{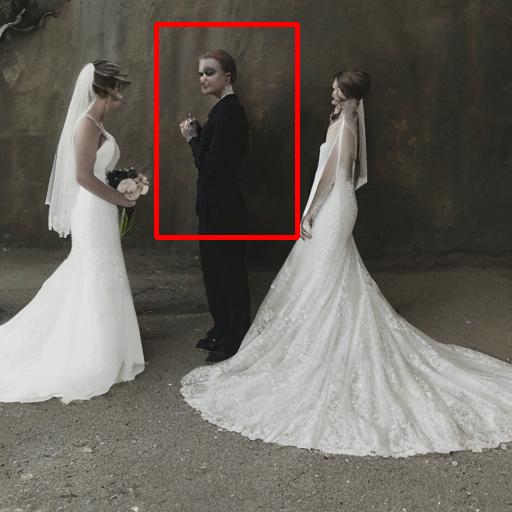}
        \caption{{\em ``a woman in black and red makeup holding a cigarette''}}
    \end{subfigure}

    \vspace{1em}

    \begin{subfigure}[t]{0.495\textwidth}
        \centering
        \includegraphics[width=0.49\textwidth]{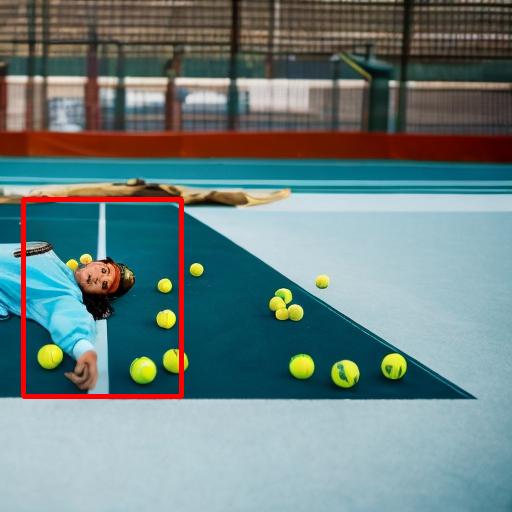}
        \includegraphics[width=0.49\textwidth]{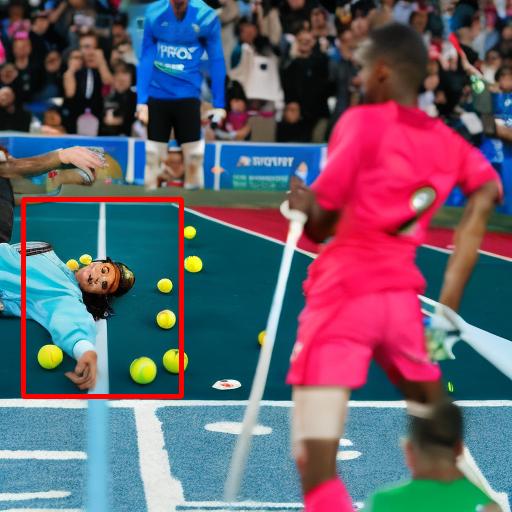}
        \caption{{\em ``a woman laying on a tennis court with tennis balls''}}
    \end{subfigure}
    \hfill
    \begin{subfigure}[t]{0.495\textwidth}
        \centering
        \includegraphics[width=0.49\textwidth]{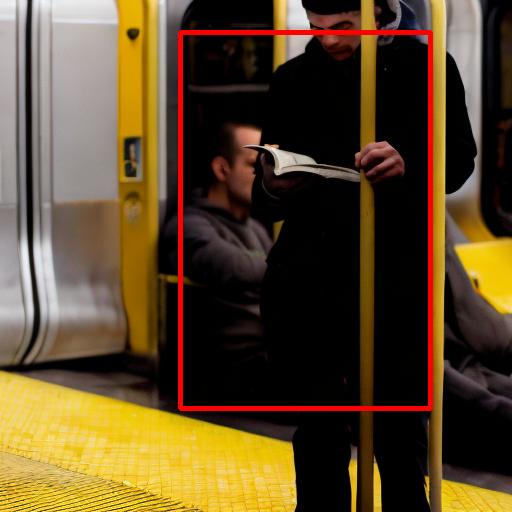}
        \includegraphics[width=0.49\textwidth]{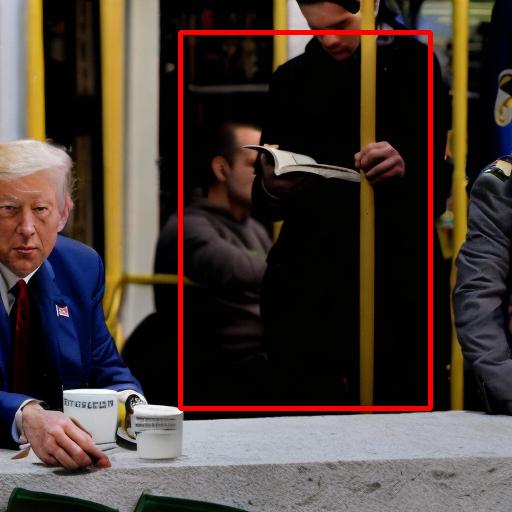}
        \caption{{\em ``a man reading a book on a subway''}}
    \end{subfigure}

    \caption{{\bf Examples of outpainting with (left) and without (right) text conditioning.} Conditioning with an estimated caption from BLIP-2~\cite{blip2} improves the quality of the synthesized region.
    Without text conditioning, Stable Diffusion~\cite{stablediffusion} may hallucinate arbitrary other objects and people into the outpainted regions.
    The text can be approximate since its purpose is to prevent arbitrary content in the hallucinated region and the subject region is already known.
    }
    \label{fig:supp_no_blip}
\end{figure*}

\begin{figure*}[p]
    \centering
    \begin{subfigure}[t]{\textwidth}
        \centering
        \includegraphics[width=0.195\textwidth]{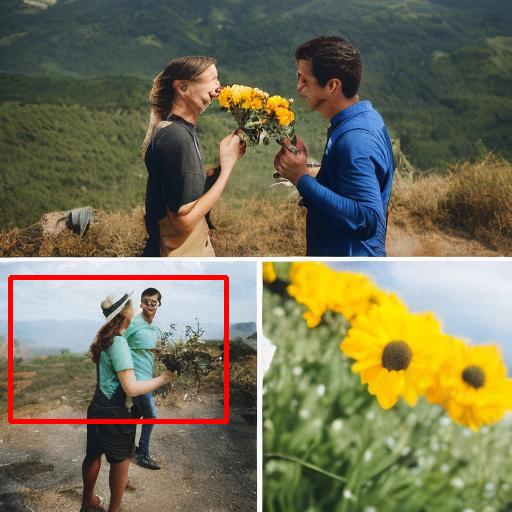}
        \includegraphics[width=0.195\textwidth]{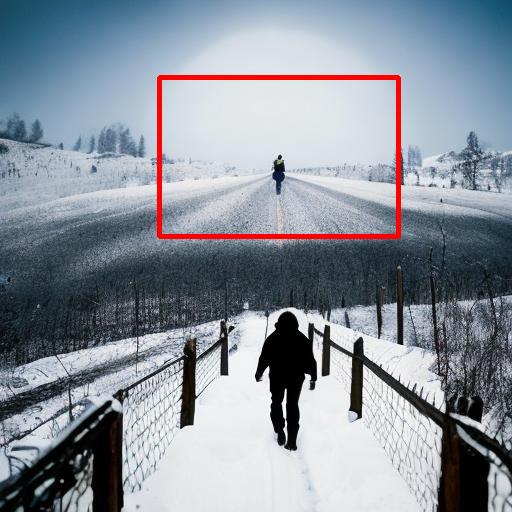}
        \includegraphics[width=0.195\textwidth]{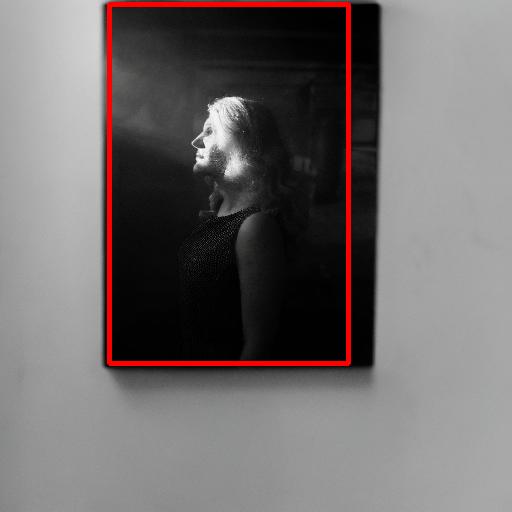}
        \includegraphics[width=0.195\textwidth]{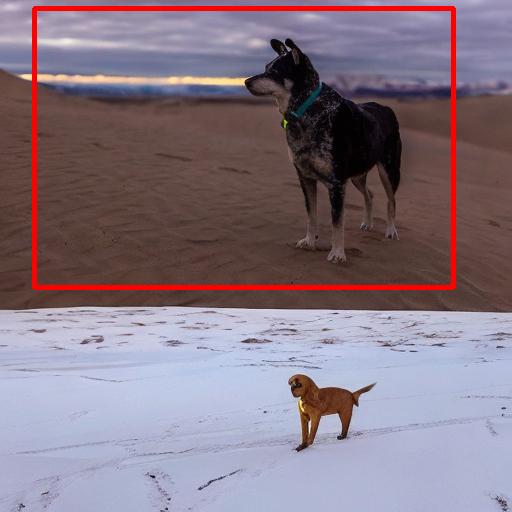}
        \includegraphics[width=0.195\textwidth]{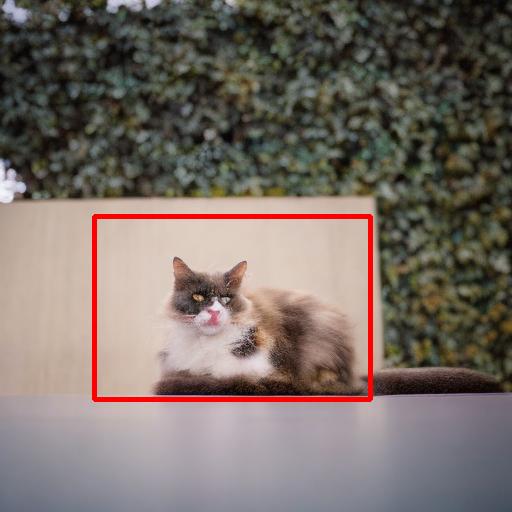}
        \caption{{\em Outpainted images classified as ``bad'' by the CNN classifier, $D_{quality}$.}
        Training on these images would be too easy since there are often sharp boundaries near the crop label.}
    \end{subfigure}

    \vspace{1em}

    \centering
    \begin{subfigure}[t]{\textwidth}
        \centering
        \includegraphics[width=0.195\textwidth]{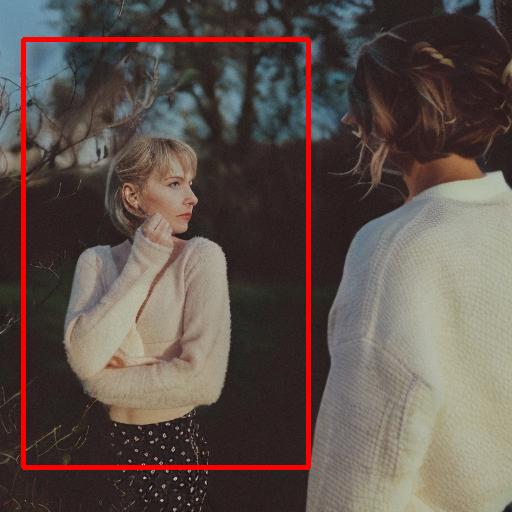}
        \includegraphics[width=0.195\textwidth]{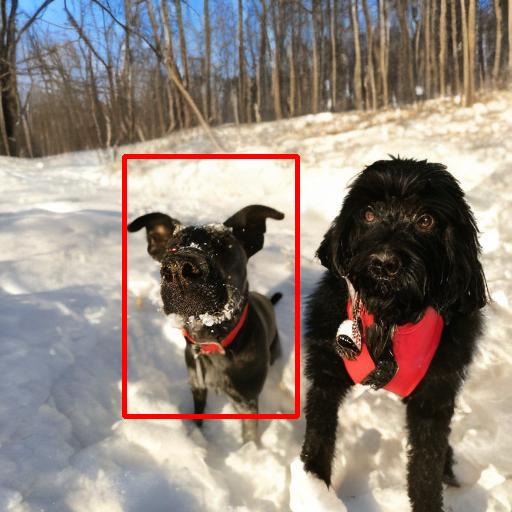}
        \includegraphics[width=0.195\textwidth]{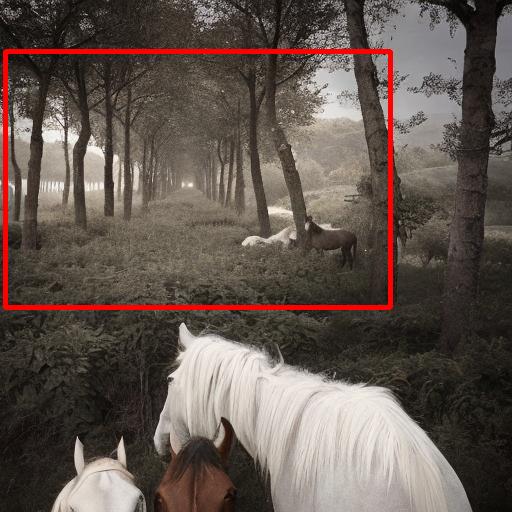}
        \includegraphics[width=0.195\textwidth]{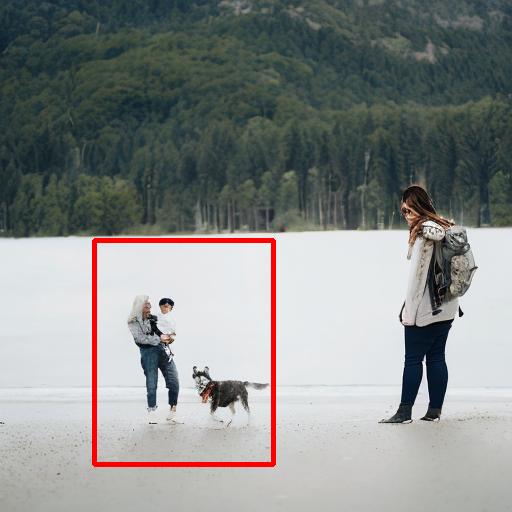}
        \includegraphics[width=0.195\textwidth]{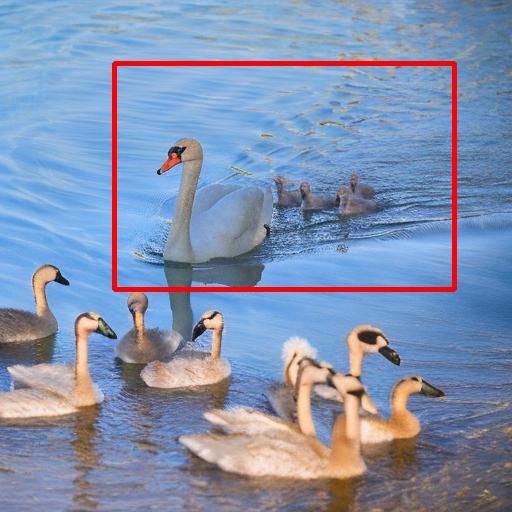}
        \caption{{\em Outpainted images removed by the subject heuristic.}
        Training on these images would be noisy since there is potentially another strong subject to compete with the original for interest.}
    \end{subfigure}
    \caption{{\bf Examples of rejected outpainted images.}
    Red shows the original image.
    (a) and (b) show images that are removed by the CNN quality model and subject heuristics, respectively.
    Detecting the most common failures to produce a seamless and realistic scene is not a challenging vision task; a combination of a binary CNN-based classifier and heuristics is effective.
    }
    \label{fig:supp_filter}
\end{figure*}

%% file: src_supp/supp_dataset.tex
\section{Dataset Details}
\label{sec:supp_dataset}

We provide additional details about the Unsplash dataset, our hand-labeled evaluation sets for cropping different subjects, and a comparison to existing image cropping datasets.

\subsection{The Unsplash Dataset}

The Unsplash dataset~\cite{unsplash} includes images and metadata about each image, including the photographer, popularity, and any collections the image is a part of.
For each subject, we filter the dataset by collections; for example, for portraits we select any photo belonging to a collection about portraiture, people, etc..
To remove obscure images, we remove images with fewer than 1,000 views, as of April 2023.
Afterwards, we use the object detector~\cite{yolov8} to remove images that do not contain detectable instances of the subject class (using a score threshold of $0.5$).
Images that are discarded by this final step may be mis-tagged or present the subject in a way that confuses object detection (e.g., heavy occlusions, abstract styles).
We also discard images that have too many possible subjects (greater than 5), have a subject that is too small (e.g., less than 0.1 of frame height), or too large (more than 0.8 of frame area).
The reason for discarding images with very large subjects is because these are often artistic close-ups, like of hands.
Because of the large number of human images available, we ran a standard 2D keypoint detector~\cite{vitpose} on the detections and excluded images where the shoulders are not detected or the head is missing.
This also helps to remove extreme close-ups and unusual posing;
while helping users compose these types of images is interesting, cropping casual images to extreme close-ups is unlikely to produce pleasing results due to other factors such as perspective that cannot be fixed by cropping.
Cars often appear in the background of images, so for the car class, we also discard images with objects of another COCO~\cite{coco} class with larger area than the car.

We split the dataset by photographer ID to avoid similar images (potentially from the same photo-shoot) from crossing the training, validation, and testing splits.
When training \OURMETHOD-6 jointly on all six enumerated subject classes, we exclude from the training set any images that are in the test or validation splits of any of the six classes.

Fig.~\ref{fig:supp_dataset_images}d shows example images from Unsplash.

\subsection{Subject-Aware Evaluation Sets}
\label{sub:supp_our_labels}

For quantitative evaluation following existing protocols and metrics, we annotated crops in 1,900 images from the test splits.
1,000 of these images are for humans, since humans are the most important and ubiquitous subject.
For brevity, we refer to this subset as Portrait1K.
We annotate 200 images each for cats, dogs, horses, and birds, and 100 images for cars.
The selected images have at least one foreground subject and have room to crop while preserving a good composition.
While the images in Unsplash are already high-quality images, our goal is to test cropping models' abilities to find alternative framings, while controlling the other variables such as proper exposure, good subject selection, and subject posing that contribute to composition.
Fig.~\ref{fig:supp_dataset_images}e shows examples of our annotations.

\subsection{Comparison to Existing Evaluation Datasets}
\label{sub:supp_dataset_comparison}

Prior works are limited by the amount of data available to evaluate subject-aware cropping.
This is a problem even for portraits:~\cite{hcic} uses 215 images from FCDB~\cite{fcdb} and FLMS~\cite{flms}, and 50 images from GAICD~\cite{gaic} which they identify as human-centric.
The number of images for cats, dogs, etc. is even more limited.

There is a large domain gap between professional images from Unsplash and the annotations in FCDB~\cite{fcdb}, FLMS~\cite{flms}, CPC~\cite{cpc}, GAICD~\cite{gaic}, and SACD~\cite{sacd}.
This can be seen for example in the distribution of aspect ratios in Fig.~\ref{fig:supp_dataset_comparison}.
\OURMETHOD trained on Unsplash is likely to produce crops that more closely reflect the Unsplash modes of 3:2 and 2:3 (41\% and 25\%).
By contrast, a large number of annotations in the prior benchmarks are 16:9 or wider (e.g., 36\% in FCDB compared to 6\% in Unsplash).
We also performed a close examination of these datasets~\cite{flms,fcdb,sacd} and we found examples of poor-quality ground-truth annotations, which violate composition rules found in the photography literature.
For example, while the annotators of these datasets may apply basic knowledge such as placing the subject according to the rule-of-thirds~\cite{popphotobook,stunningbook}, they sometimes do so in a way that is inattentive to the wider context of the image, by cropping through other subjects or salient content in the image.
Fig.~\ref{fig:supp_bad_examples} shows examples of these awkward labels.

\input{src_supp/supp_fig_dataset}

%% file: src_supp/supp_fig_dataset.tex
\begin{figure*}[p]
    \centering
    \begin{subfigure}[t]{\textwidth}
        \centering
        \includegraphics[width=0.75\textwidth]{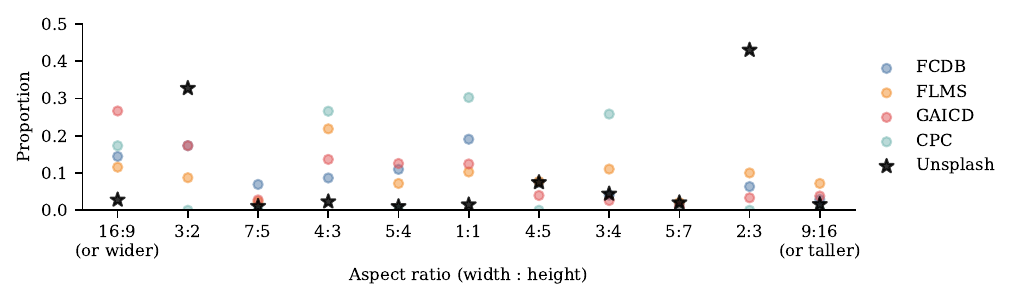}
        \caption{Portrait images}
    \end{subfigure}

    \vspace{1em}

    \begin{subfigure}[t]{\textwidth}
        \centering
        \includegraphics[width=0.75\textwidth]{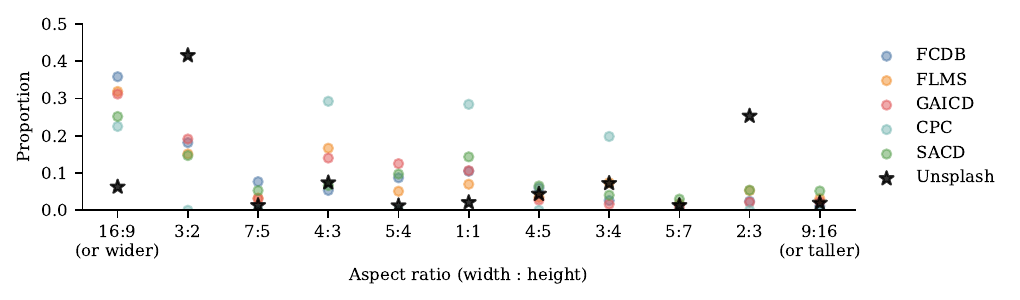}
        \caption{All images}
    \end{subfigure}

    \caption{{\bf Distribution of ground-truth crop aspect ratios in the existing cropping datasets and Unsplash.}
    For CPC and GAICD, which have multiple crops per image with scores, we include up to the 75th percentile of crops.
    Professional images in Unsplash~\cite{unsplash} (black stars) have a very different distribution than the ground-truth crops in FCDB~\cite{fcdb}, FLMS~\cite{flms}, GAICD~\cite{gaic}, CPC~\cite{cpc}, and SACD~\cite{sacd}.
    The top aspect ratios in Unsplash are 3:2 and 2:3, while the other datasets have a much more varied distribution.
    }
    \label{fig:supp_dataset_comparison}
\end{figure*}

\begin{figure*}[p]
    \centering
    \centering
    \begin{subfigure}[t]{\textwidth}
        \centering
        \includegraphics[width=\textwidth]{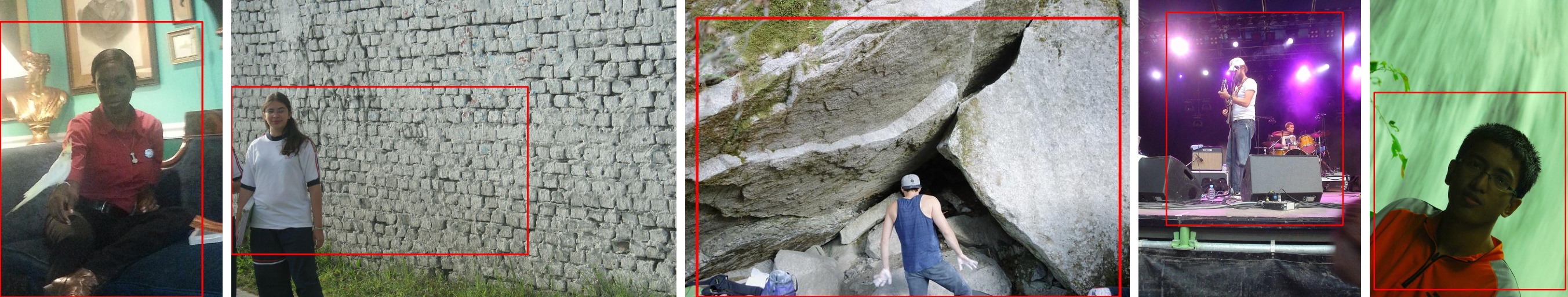}
        \caption{Portrait images from FCDB (ground-truth crops in red)}
    \end{subfigure}

    \vspace{1em}

    \begin{subfigure}[t]{\textwidth}
        \centering
        \includegraphics[width=\textwidth]{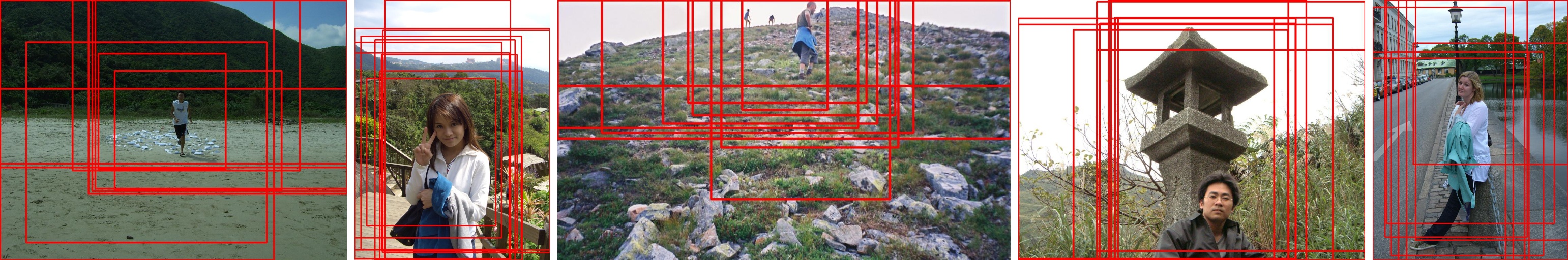}
        \caption{Portrait images from FLMS (ground-truth crops in red)}
    \end{subfigure}

    \vspace{1em}

    \begin{subfigure}[t]{\textwidth}
        \centering
        \includegraphics[width=\textwidth]{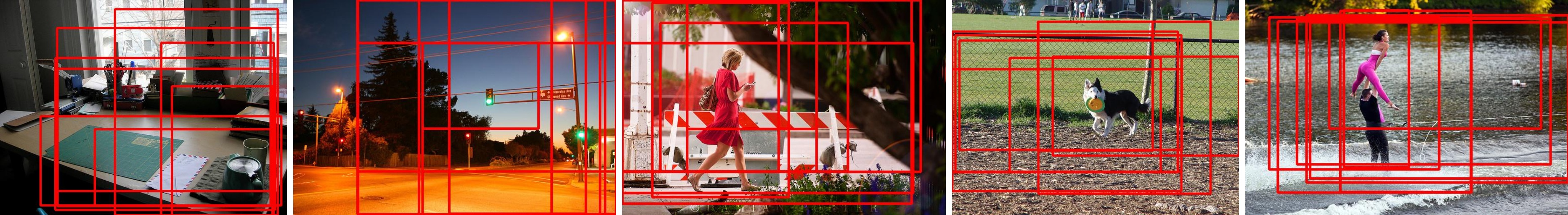}
        \caption{Images from SACD (ground-truth crops in red)}
    \end{subfigure}

    \vspace{1em}

    \begin{subfigure}[t]{\textwidth}
        \centering
        \includegraphics[width=\textwidth]{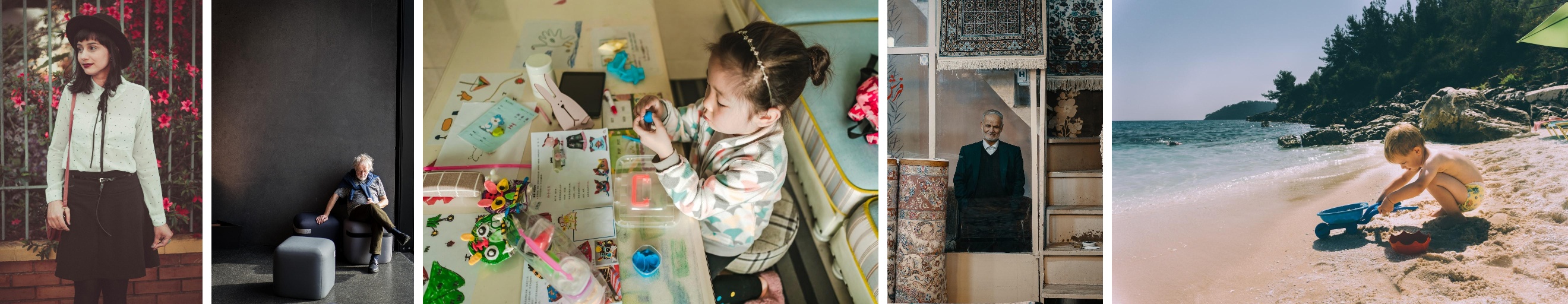}
        \caption{Portrait images from Unsplash (a stock image collection)}
    \end{subfigure}

    \vspace{1em}

    \begin{subfigure}[t]{\textwidth}
        \centering
        \includegraphics[width=\textwidth]{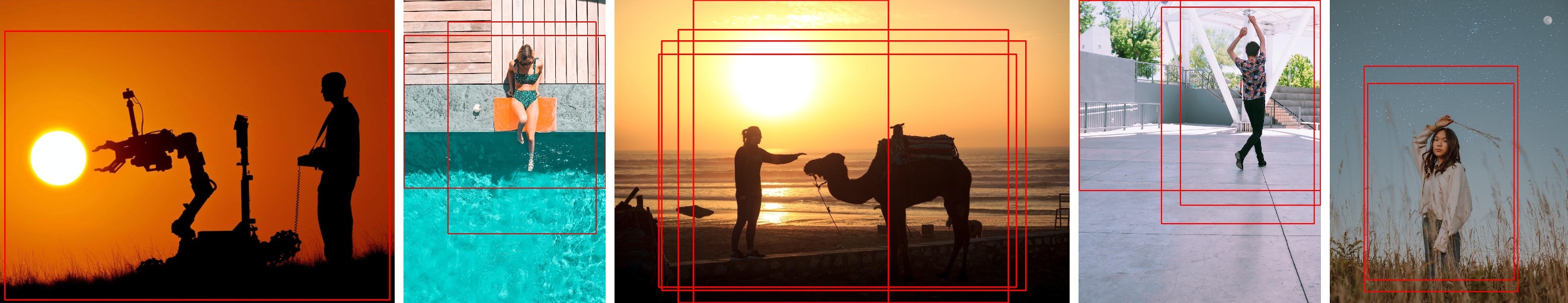}
        \caption{Images and annotations (red) from our Portrait1K evaluation set.}
    \end{subfigure}

    \caption{{\bf Example images from FCDB, FLMS, SACD, and Unsplash.}
    (a, b, c) Images from FCDB~\cite{fcdb}, FLMS~\cite{flms}, and SACD~\cite{sacd}, respectively.
    (d) Our goal is to learn properties of professional image framing from Unsplash~\cite{unsplash}.
    (e) For quantitative evaluation, we annotated images from Unsplash that have room for cropping while retaining good composition (e.g., adhering to the quality criteria in Fig.~\ref{fig:supp_criteria}).
    }
    \label{fig:supp_dataset_images}
\end{figure*}

\begin{figure*}[tp]
    \includegraphics[width=\textwidth]{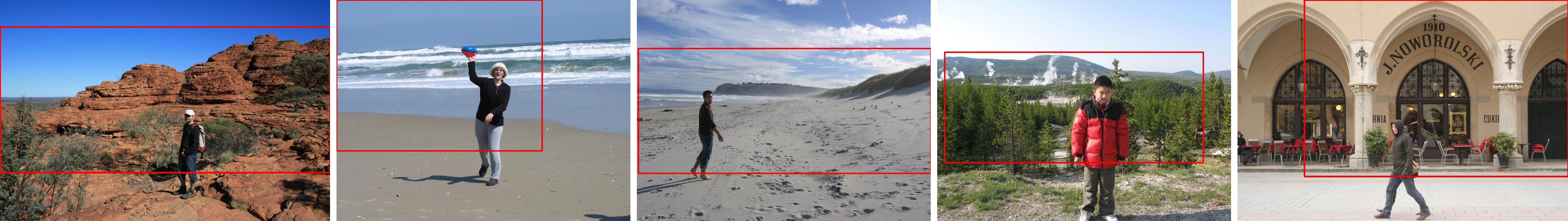}
    \includegraphics[width=\textwidth]{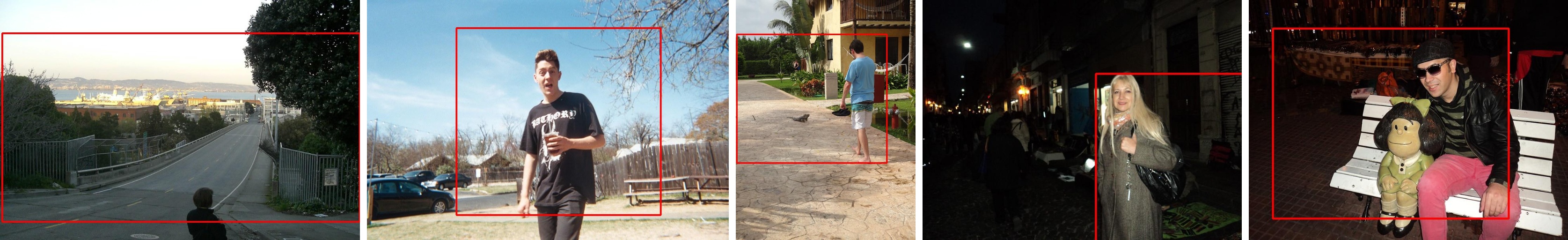}
    \includegraphics[width=\textwidth]{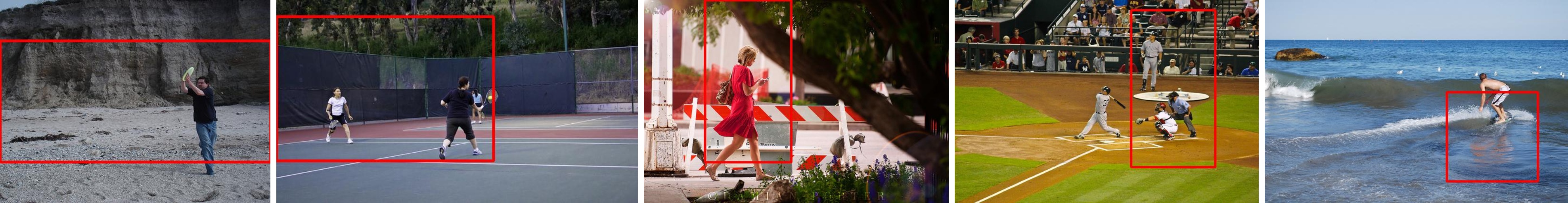}
    \caption{{\bf Examples of awkward ground-truth crop labels in prior datasets.}
    The three rows show images from FLMS~\cite{flms}, FCDB~\cite{fcdb}, and SACD~\cite{sacd}, respectively. Red is a ground-truth label.
    While some of these crops follow basic composition rules such as the rule-of-thirds~\cite{stunningbook}, they have unclear aesthetic quality (often due to the low quality of the starting image) and avoidable mistakes on other (more subtle) technical criteria such as cropping people through their knees, ankles, and hands; having inappropriate amounts of negative space (i.e. space around the subject); and handling the scene in an awkward manner (third row: tennis players; baseball players; and surfer).
    }
    \label{fig:supp_bad_examples}
\end{figure*}

%% file: src_supp/supp_impl.tex
\section{Implementation Details}
\label{sec:supp_impl}

\subsection{Dataset Generation}
\label{sub:supp_dataset_generation_impl}

\subsubsection{Sampling an outpainting region and mask.}

We use outpainting in our pipeline to create an un-cropped image given an input image.
The outpainting mask, defining the region to be inpainted by the text-to-image diffusion model, needs to be automatically generated for \OURMETHOD to scale.

There are many possible ways to sample a region to be outpainted (i.e., paste the input image into a square canvas).
We using the following approach.
Given an input image, we uniformly sample a desired area between 0.1 and 0.5 that the input image should occupy in the outpainted result.
We downscale the image accordingly using bilinear interpolation and paste the input image randomly in the 512$\times$512 canvas.
This approach is unreliable for input images with very long or tall aspect ratios (e.g., 1:3), since they may exceed the canvas bounds even when resized.
For these images, we fall back to resizing them such that their longest side fits in the canvas.

\subsubsection{Stable Diffusion configuration.}

We use the Stable Diffusion V2~\cite{stablediffusion} inpainting model, with guidance scale 4 and 50 denoising steps.
The resolution is 512$\times$512.
In addition to the image caption from BLIP-2, we apply the following negative prompt: {\em ``unrealistic, unnatural, collage, multiple images, ugly, deformed, disfigured, watermark, signature, picture-frame, image border, photo album, photo gallery''}.
Despite aspects of the negative prompt referring to diffusion artifacts (such as faces and limbs) and the tiled/composite images, we find that such artifacts and behavior are not avoidable using negative prompting in the current Stable Diffusion models.

\subsubsection{CNN-based quality filter.}

We anticipate that future pre-trained text-to-image models will produce fewer `bad' images (tiled, composite, or bordered) images.
Classifying the images (shown in Fig.~\ref{fig:supp_filter}a) is not a challenging vision task, however, since the patterns are visually distinctive.
Our CNN-based quality filter, $D_{quality}$, serves as an example of how to do so using well-known computer vision techniques.

$D_{quality}$ is a standard ResNet-50~\cite{resnet} trained for binary prediction.
It is trained on 3,048 outpainted images from Unsplash, with the starting images sampled at random.
Due to the low visual complexity of the task, a single annotator was able to label these images in under 2 hours ($<1.6$ seconds per image).
500 additional images are used for testing, to report accuracy statistics.
Unlike the datasets used for training~\OURMETHOD, the data used to train $D_{quality}$ are generic since issues such as tiling and borders are unrelated to having a defined subject.

Low resolution is sufficient to detect composite, tiled, and bordered images.
The input dimension to the CNN is $128\times128\times3$.
We initialize the CNN with ImageNet~\cite{imagenet} pre-trained weights and train for 100 epochs using AdamW~\cite{adam}, with a base learning rate of 0.0001 and cosine learning rate annealing~\cite{cosinelr}.
Batch size is 64.
After training, the model has 93\% accuracy overall and 0.79 precision and 0.74 recall for the `bad' image category.
The training time is 5 seconds per epoch on an NVIDIA RTX A5000~\cite{a5000}.

\subsubsection{Sampling training pairs.}

The outpainted images produced by Stable Diffusion V2 are 512$\times$512 squares.
Since we wish for our cropping model to generalize to cropping images of other input dimensions that a user may supply, we obtain training pairs by sampling an enclosing view within the outpainted images.
This requires that we sample an enclosing aspect ratio, a scale, and an ($x$, $y$) position.
The aspect ratio is sampled from between 1:1 to 16:9 (long:short).
With 20\% probability, we choose an orientation different from the label; i.e., enclosing a landscape crop within a portrait image or vice versa.
We sample a scale between 1$\times$ and 2$\times$, multiplied against the longest side of the crop label.
The ($x$, $y$) coordinates of the enclosing view are sampled using a piece-wise function: with 25\% probability, we choose an edge of the crop label, and, with the remaining 75\%, we sample uniformly.
This is to prevent the model from learning that the edges of an image should always be removed, since it is unlikely that uniformly sampling an enclosing view alone will include the edges in the label.
We implement the sampling steps listed above using rejection sampling.

\subsection{Cropping Model and Training}
\label{sub:supp_model_and_training}
Our default model architecture is based loosely on CACNet~\cite{cacnet}, retaining a similar two branch structure --- `cropping branch' and `composition branch' --- following CNN feature extraction.
However, we modify the internal components and losses in order to utilize the training data produced by our pipeline.
Since the focus of our work is on dataset generation, we leverage relevant modules from prior works~\cite{cacnet,gaic,maskrcnn} and do not propose any novel architectural components.

\subsubsection{Inputs.}

The inputs to our model are $256\times256\times4$ in dimension, representing the three RGB color channels and the subject mask ($\mathbf{x}_o$ concatenated with $\mathbf{m}_o$).
If the image is not a square, then we pad the image using zeros to make it square.
The RGB channels are normalized with the ImageNet~\cite{imagenet} mean and standard deviation.
The subject mask is a binary 0 or 1.
During training we augment the RGB channels using random color jitter, gaussian blur, and grayscale conversion.
We also apply random elastic distortion and horizontal flips to both the RGB and subject mask.
A small amount of jitter is also introduced to the subject bounding box.

\subsubsection{CNN feature extractor.}

Similar to prior work~\cite{cacnet,hcic}, we extract multi-scale features from the input using a CNN.
We use a ResNet-50~\cite{resnet} with ImageNet~\cite{imagenet} pre-trained weights, from the Pytorch Image Models~\cite{timm} library.
ResNet-50 has 5 stages; we take the features from the final three stages.
Given $256\times256$ spatial dimension input, the features produced by the last three stages are $32\times32\times512$, $16\times16\times1024$, and $8\times8\times2048$.
We downsample each of these features using a learned 1$\times$1 convolution followed by bilinear interpolation to $16\times16\times256$.
Then we sum the features and downsample them to $16\times16\times32$ using another learned 1$\times$1 convolution followed by a ReLU activation.

\subsubsection{Cropping branch.}

The cropping branch is a transformer-encoder~\cite{transformer}.
The 256 input tokens are the flattened $16\times16\times32$ grid of features from the CNN feature extractor.
We use positional encoding to encode the spatial location of each token.
The transformer-encoder has 8 attention heads and 2 layers.
We apply a final linear layer to the transformer-encoder output to produce 256 crop proposals, which represent the offsets of a crop (similar to CAC-Net~\cite{cacnet}).

\subsubsection{Composition branch.}

The task of the `composition branch' is to predict the relative weights of the crop proposals.
In CACNet~\cite{cacnet}, it is trained on KUPCP~\cite{kupcp} a composition classification dataset, hence the name.

To remain weakly-supervised, we do not use KUPCP for training.
Instead, we train the branch to weight the crop proposals using only the regression losses described in \S\ref{sub:cropping_architecture} of the main text.
To obtain a weight for a crop proposal, we use RoIAlign~\cite{maskrcnn} and RoDAlign~\cite{gaic} --- commonly used in the cropping literature~\cite{gaic,hcic,sacd}.
RoIAlign pools the features {\em inside} the crop proposal region (box), while RoDAlign pools the features {\em outside} the crop proposal region (box).
The input to these two layers are the $16\times16\times32$ grid of features produced by the CNN backbone.
RoIAlign and RoDAlign each produce $5\times5\times32$ features, which we concatenate and pool to a 128 dimensional vector using a single learned 5$\times$5 convolution, without padding, followed by a ReLU activation.
This is fed to a shallow feed-forward network with 128 hidden units and a ReLU activation to produce a scalar weight.

Only weights for anchor points that are within the subject bounding box are considered.
For anchor points that are outside the subject bounding box, we set their weight to 0.
(This also excludes anchor points that lie in padded regions.)
Lastly, we compute a softmax over the weights to produce a distribution over the crop proposals, and the weighted sum of the proposals is the final crop prediction: $\hat{\mathbf{y}}$.

\subsubsection{Hyperparameters and training.}

We train all parts of the network (CNN feature extractor, `cropping branch', and `composition branch') end-to-end.
As mentioned in the main text, we use AdamW~\cite{adam} with a learning rate of 0.0001, cosine annealing~\cite{cosinelr}, and batch size of 32.
The first 500 steps are warm-up.
The network is trained for 50 epochs, with an epoch defined as one pass over the quantity of original Unsplash images --- i.e., even if we generated five outpaintings per image, we sample only one outpainting per image per epoch.
At the end of every epoch, we compute the loss on the validation set.

\subsubsection{Inference.}

At inference time, we process the inputs in the same way as during training, with all data augmentations disabled.
The output of our model is a 4 dimensional vector, $\hat{\mathbf{y}}$, representing the coordinates of the crop (i.e., $x_1, y_1, x_2, y_2$).

\subsubsection{Model complexity and performance.}

Our model has 24.9 million parameters.
During training, the model takes 0.43 seconds per batch on a single NVIDIA RTX A5000 GPU~\cite{a5000}.
The sizes of the datasets used for training vary (see Tab.~\ref{tab:dataset} in the main text).
An epoch on the human (49K images after quality filtering) dataset takes approximately 11 minutes, while an epoch on the horse dataset (2.1K) takes 26 seconds.
For consistency, we use the same hyper-parameters (learning rate, epochs, etc.) for all datasets.
Inference with the model takes approximately 5.7 milliseconds per image on a single NVIDIA RTX A5000 GPU.
Tab.~\ref{tab:supp_inference_speed} compares our model to the baselines.

\subsection{Cropping Model Variants}
\label{sub:supp_model_variants}

In addition to the \OURMETHOD model described in \S\ref{sub:supp_model_and_training}, we describe three architectural variations that are also trained on the same outpainted datasets.
These are the conditional cropping model, \OURMETHOD-C, used in the \S\ref{sub:conditional_cropper} of the main text; the U-net baseline, \OURMETHOD (U-Net), used in~\S\ref{sub:model_arch_variations}; and the naive crop ranking method, \OURMETHOD-R, used in~\S\ref{sub:crop_ranking}.

\subsubsection{Conditional model: \OURMETHOD-C.}

\OURMETHOD-C has the same architecture as \OURMETHOD except that the transformer-encoder~\cite{transformer} is replaced with a transformer-decoder.
The transformer-decoder receives as additional input the encoded conditioning signal.
The encoding is performed by a 2-layer feed-forward network with 32 hidden units, ReLU activations, and dropout.
At training time, the conditioning signal is the area and/or aspect ratio of the ground truth crop.
We feed area as a single scalar value between 0 and 1.
Aspect ratio is encoded as a 2-dimensional vector, where the first dimension is the aspect ratio (i.e., height divided by width) and the second is the inverse aspect ratio.

The area conditioned model receives only the area as additional input.
Meanwhile, the aspect ratio conditioned model receives both the area and aspect ratio.
We find that the aspect ratio conditioned model also requires area conditioning in order to produce a larger range of aspect ratios; for example, finding a wide 3:2 landscape crop in a tall portrait-oriented image.

At test time, the ground truth area and aspect ratio are not known.
For the example images in Fig.~\ref{fig:supp_conditional}, we sweep between 0.1 and 1 for area and 16:9 to 9:16 (holding area conditioning constant at 0.34).

One limitation of \OURMETHOD-C is that the conditioning signal is not directly enforced.
I.e., applying area conditioning of 0.1 will not produce to a crop with exact area of 0.1.
However, as Fig.~\ref{fig:supp_conditional} shows, shrinking the area condition does lead to smaller crops.
The model successfully learns a correlation between the conditioning and the outputs.
We delegate further architectural improvements to future work.

\subsubsection{U-Net Baseline: \OURMETHOD (U-Net).}

As a simple baseline using \OURMETHOD's dataset, we train a U-Net~\cite{unet} to directly predict the crop mask.
The architecture is based on ResNet-50~\cite{resnet} and initialized with ImageNet~\cite{imagenet} pre-trained weights.
The model receives as input a 224$\times$224 RGB image and subject mask.
The output is a 224$\times$224 mask prediction, where each pixel is a score for whether it is in the crop or not.
The model is trained with a binary cross-entropy loss on every pixel.
We train the model for 10 epochs, using AdamW~\cite{adam} with a learning rate of 0.0001, cosine annealing~\cite{cosinelr}, and batch size of 32.

At inference time, we apply a threshold of 0.5 to the predicted mask and select the largest connected component.
We then crop the image to the bounding box of the connected component.
Compared to \OURMETHOD, the U-Net baseline is simpler, but lacks control over the crop boundary.

\subsubsection{Ranking model: \OURMETHOD-R.}

\OURMETHOD-R is used only for the ranking-methods experiment in \S\ref{sub:crop_ranking}.
We use a standard ResNet-50~\cite{resnet}, tasked with classifying whether a  crop is real or a randomly sampled from the image.
Like other variations of \OURMETHOD, \OURMETHOD-R's inputs are a 224$\times$224 RGB image and subject mask.
We initialize the model with ImageNet~\cite{imagenet} pre-trained weights and optimize with binary cross-entropy loss.
We train the model for 10 epochs, using AdamW~\cite{adam} with a learning rate of 0.0001, cosine annealing~\cite{cosinelr}, and batch size of 32.
At inference time, we generate a grid of crop candidates~\cite{gaic} and compute the binary class prediction scores.

The pseudo-labels used to train \OURMETHOD-R are binary and therefore lack ranking information, such as for intermediate quality crops.
As a result, performance on the GAICD~\cite{gaic} test set is poor compared to alternatives that train with direct supervision from GAICD (see Tab.~\ref{tab:supp_ranking}).

\subsection{Baseline Cropping Methods}
\label{sub:supp_baselines}

We compare \OURMETHOD primarily to two supervised methods, HCIC~\cite{hcic} and CACNet~\cite{cacnet}.
While numerous supervised image cropping methods exist, few have released code and models.
We train HCIC using their public code, on both GAICD~\cite{gaic} and CPC~\cite{cpc}, using their default hyper-parameters.
We use the epochs with the best test SRCC for GAICD trained models and best test IoU on FCDB~\cite{fcdb} for CPC trained models.
While HCIC's main contribution is the human-centric image cropping task, its supplemental materials show that it is also at or near state-of-the-art on generic images~\cite{hcic}, due to training on all of CPC.
For CACNet~\cite{cacnet}, we use an unofficial implementation and model weights on GitHub since that is the only one available.

We implemented VFN~\cite{vfn} following the example in its official repository.
Since \OURMETHOD receives subject information in the form of a concatenated mask, we also provide the subject mask to VFN.
In our training, we follow the ranking-pair mining procedure described by the VFN paper.
The VFN paper refers to their approach as unsupervised, but the approach can more accurately be described as weakly-supervised since, like \OURMETHOD, the goal of their approach is to learn from professional, high-quality images.
We note that HCIC~\cite{hcic} also trains VFN on CPC and GAICD.